\documentclass[10pt,twocolumn,letterpaper]{article}

\usepackage{cvpr}              

\usepackage{bm} 
\usepackage{times}
\usepackage{epsfig}
\usepackage{graphicx}
\usepackage{amsmath}
\usepackage{amsthm}
\usepackage{amssymb}
\usepackage{amsfonts}
\usepackage{url}
\usepackage{algorithm}
\usepackage{algorithmicx}
\usepackage{algpseudocode}
\usepackage{booktabs}
\usepackage{threeparttable}
\usepackage{multirow}
\usepackage{subcaption}
\usepackage{caption}
\usepackage{array}
\usepackage{xcolor}
\usepackage{soul}
\usepackage{footnote}
\usepackage{tablefootnote}
\usepackage{bbding}
\usepackage{diagbox}
\usepackage[pagebackref,breaklinks,colorlinks]{hyperref}

\newcommand{\lightgray}[1]{{\color{lightgray}{#1}}}

\newcommand{\tabincell}[2]{\begin{tabular}{@{}#1@{}}#2\end{tabular}}
\DeclareMathOperator*{\argmax}{argmax}
\DeclareMathOperator*{\argmin}{argmin}

\renewcommand{\a}{{\bm{a}}}

\renewcommand{\c}{{\bm{c}}}

\renewcommand{\r}{{\bm{r}}}
\newcommand{\s}{{\bm{s}}}

\newcommand{\balpha}{{\bm{\alpha}}}

\newcommand{\qileft}{[\kern-0.15em[}
\newcommand{\qiLeft}{\left[\kern-0.4em\left[}
\newcommand{\qiright}{]\kern-0.15em]}
\newcommand{\qiRight}{\right]\kern-0.4em\right]}
\newcommand{\A}{{\mathcal{A}}}

\newcommand{\C}{\mathcal{C}}

\newcommand{\N}{\mathcal{N}}

\newcommand{\W}{\mathcal{W}}
\usepackage[capitalize]{cleveref}
\crefname{section}{Sec.}{Secs.}
\Crefname{section}{Section}{Sections}
\Crefname{table}{Table}{Tables}
\crefname{table}{Tab.}{Tabs.}

\begin{document}

\title{ViTAS: Vision Transformer Architecture Search}

\author{Xiu Su$^{1}$ \quad Shan You$^{2,3}$\thanks{Correspondence to: Shan You $<$\texttt{youshan@sensetime.com}$>$} \quad Jiyang Xie$^{4}$ \quad Mingkai Zheng$^{1}$ \quad Fei Wang$^5$ \\
Chen Qian$^2$ \quad Changshui Zhang$^{3}$ \quad Xiaogang Wang$^{2,6}$ \quad Chang Xu$^{1}$ \\
~\\
$^1$School of Computer Science, Faculty of Engineering, The University of Sydney\\
$^2$SenseTime Research \\
$^3$Department of Automation, THUAI, BNRist, Tsinghua University\\
$^4$Beijing University of Posts and Telecommunications \\
$^5$University of Science and Technology of China\\
$^6$The Chinese University of Hong Kong
}

\maketitle

\newtheorem{theorem}{Condition}

\begin{abstract}
Vision transformers (ViTs) inherited the success of NLP but their structures have not been sufficiently investigated and optimized for visual tasks. One of the simplest solutions is to directly search the optimal one via the widely used neural architecture search (NAS) in CNNs. However, we empirically find this straightforward adaptation would encounter catastrophic failures and be frustratingly unstable for the training of superformer. In this paper, we argue that since ViTs mainly operate on token embeddings with little inductive bias, imbalance of channels for different architectures would worsen the weight-sharing assumption and cause the training instability as a result. Therefore, we develop a new cyclic weight-sharing mechanism for token embeddings of the ViTs, which enables each channel could more evenly contribute to all candidate architectures. Besides, we also propose identity shifting to alleviate the many-to-one issue in superformer and leverage weak augmentation and regularization techniques for more steady training empirically. Based on these, our proposed method, ViTAS, has achieved significant superiority in both DeiT- and Twins-based ViTs. For example, with only $1.4$G FLOPs budget, our searched architecture has $3.3\%$ ImageNet-$1$k accuracy than the baseline DeiT. With $3.0$G FLOPs, our results achieve $82.0\%$ accuracy on ImageNet-$1$k, and $45.9\%$ mAP on COCO$2017$ which is $2.4\%$ superior than other ViTs.

\end{abstract}
\vspace{-4mm}
\section{Introduction}
\label{sec:intro}
Transformer, as a self-attention characterized neural network, has been widely leveraged for natural language processing (NLP) tasks~\cite{devlin2018bert,radford2018improving,radford2019language,brown2020language}. Amazingly, recent breakthrough of vision transformers (ViTs)~\cite{dosovitskiy2020image,touvron2020training} further revealed the huge potential of transformers in computer vision (CV) tasks. With no use of inductive biases, self-attention layers in the transformer introduce a global receptive field, which conveys refresh solutions to process vision data. Following ViTs, there have been quite a few works on vision transformers for a variety of tasks, such as image recognition~\cite{dosovitskiy2020image,touvron2020training,yuan2021tokens,han2021transformer,liu2021swin,wu2021cvt}, object detection~\cite{zhu2020deformable,liu2021swin}, and semantic segmentation~\cite{liu2021swin}.

Despite the remarkable achievements of ViTs, the design of their architectures is still rarely investigated. Current ViTs simply split an image into a sequence of patches (\ie, tokens) and stack transformer blocks as NLP tasks. Nevertheless, this vanilla protocol does not necessarily ensure the optimality for vision tasks. Stacking manner and intrinsic structure of the blocks need to be further analyzed and determined, such as patch size of input, head number in multihead self-attention (MHSA), output dimensions of parametric layers, operation type, and depth of the whole model. Therefore, we raise questions that \textit{What makes a better vision transformer? How can we obtain it?} Inspired by the success of one-shot neural architecture search (NAS) in ConvNets (CNNs), our intuition is also to directly search for an optimal architecture for ViTs, which in turn gives us insight about designing more promising ViTs.


\begin{figure*}[!t]
    \centering
    \includegraphics[height=.5\linewidth,width=1.0\linewidth]{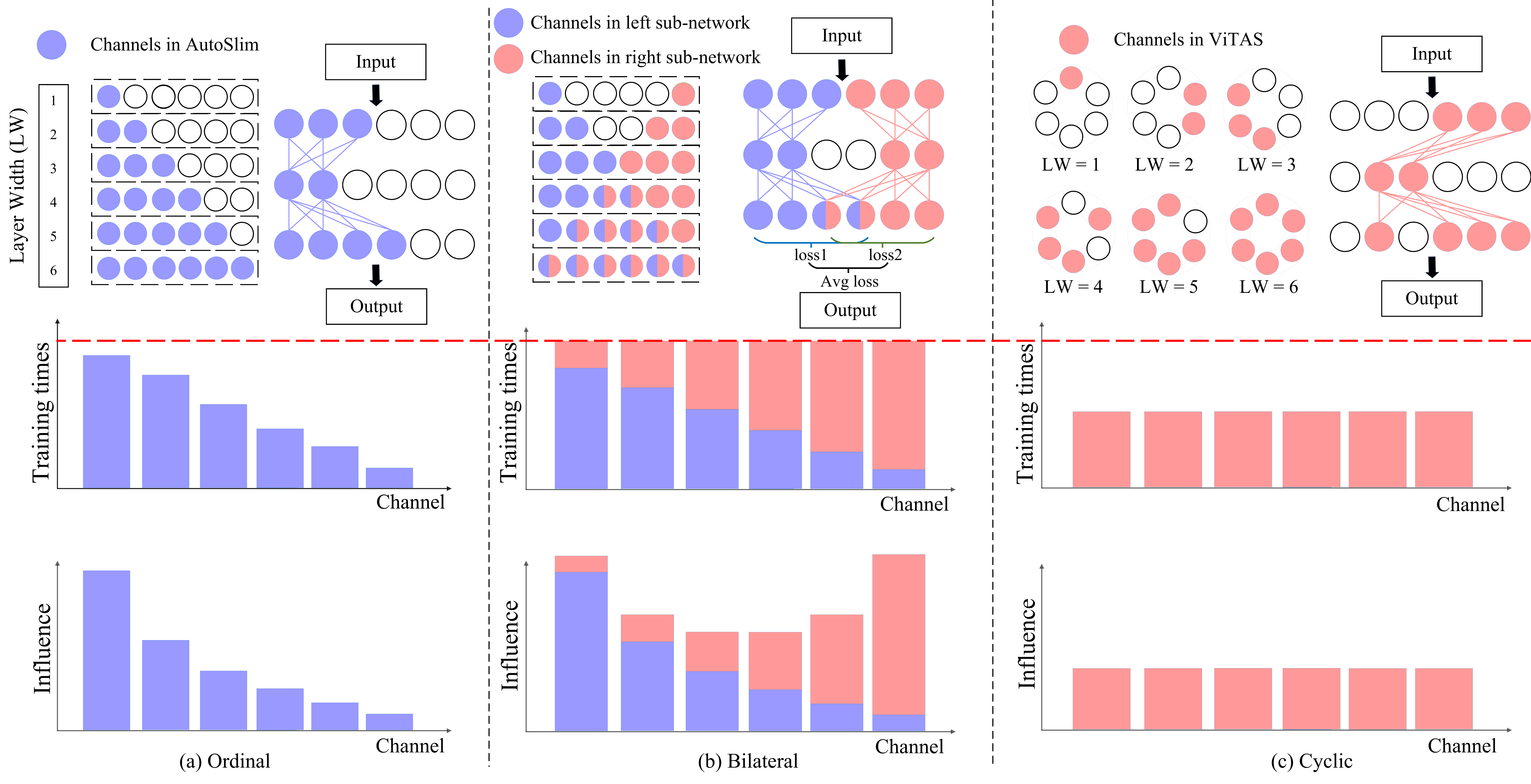}
    \caption{Comparison between (a) ordinal~\cite{autoslim}, (b) bilateral~\cite{su2021bcnet} and (c) our cyclic forms with a toy example of six groups of channels. In (a), channels cannot be fairly trained in terms of training times and influence. Channels in (b) need double training cost compared to the others and also unevenly distributed influence. Then in (c), the proposed cyclic pattern overcomes the defects of (a) and (b) and achieves fairly training of channels~\wrt~training times and influence while maintaining half training cost than the bilateral form.}\label{fig:loop}
    \vspace{-6mm}
\end{figure*}

Unlike the sliding convolutions of CNNs, ViTs project the patches into a sequence of token embeddings, and the features are extracted sequentially. In this way, how to specify an appropriate configuration (dimension) for token embeddings of all layers play an important role for the architecture of ViTs. To search for the optimal token embedding dimension, recent work~\cite{chen2021autoformer,li2021bossnas} simply borrow the ordinal weight sharing~\cite{guo2020single,autoslim} in CNNs for the superformer (a.k.a. supernet in CNNs) to accommodate different token dimension. 
However, this ordinal mechanism would inevitably introduce imbalance among channels during training, causing the superformer cannot evaluate each token dimension well and induces sub-optimal architectures consequently. Though recent bilateral mechanism~\cite{su2021bcnet} was proposed to handle this issue, the training cost has to be doubled yet the imbalance of channels still exist to some extent.

In this paper, we propose a novel \textit{cyclic} weight sharing mechanism for superformer to embody various token embedding dimensions of all layers. Concretely, we encourage balanced \textit{training fairness} and \textit{influence uniformity} for each channel in the superformer. With these two conditions, the cyclic rule could be learned as an index mapping to indicate each dimension of token embeddings (see Figure \ref{fig:loop}), so that each could be more evenly evaluated. Besides, since the cyclic rule is a single-pass mapping, computation cost of training the superformer is similar to that of a ordinal one. 

Based on the customized cyclic manner, we propose a corresponding NAS method for ViTs dubbed vision transformer architecure search (ViTAS). However, we empirically observe that the training of superformer tends to be frustratingly instable. We argue that the space size of ViTs are way too huger (even $1.1 \times 10^{54}$), and propose to calibrate the space with an identity shifting technique. Besides, we find that strong augmentation and regularization are critic to further stabilize the superformer training. Extensive experimental results have shown the superiority of our ViTAS. Our main contributions can be summarized as follows.

\begin{itemize}
    \vspace{-3mm}
    \item We introduce a novel cyclic weight sharing mechanism for token embedding dimension in superformer, which ensures an improved balance of channels and thus contributes to more accurate evaluation. 
    
    \vspace{-3mm}
    \item Based on cyclic rule, our proposed method ViTAS leverages identity shifting to calibrate the ViTs' space, and encourages weak augmentation and regularization to stabilize the training of superformer. 
    
    \vspace{-3mm}
    \item Our ViTAS has achieved state-of-the-art performance in various tasks and FLOPs levels. For example, with Twins transformer space and $3.0$G FLOPs, ViTAS achieved $82.0\%$ top-$1$ accuracy on ImageNet-$1$k, and $45.9$ ($44.4$)\% mAP with Mask RCNN (RetinaNet) framkwork on COCO$2017$ dataset, achieves $2.4\%$ ($2.2\%$) superior than other ViTs. With searched results transfer to ADE$20$K dataset, ViTAS also surpass other baselines by $2.0\%\sim4.0\%$ on mIoU. Moreover, With DeiT search space and $1.4$G FLOPs, ViTAS achieves $75.6\%$ on Top-$1$ accuracy and is $3.3\%$ than the DeiT baseline.

    \vspace{-2mm}
\end{itemize}

\section{Related Work}

\textbf{Vision Transformer.} ViT was first proposed by Dosovitskiy et al.~\cite{dosovitskiy2020image} to extend the applications of transformers into computer vision fields by cascading manually designed multilayer perceptrons (MLPs) and MHSA modules. Touvron et al.~\cite{touvron2020training} introduced a teacher-student strategy and a distillation token into the ViT, namely data-efficient image transformers (DeiT). 
Recently, other variants of ViT were proposed and all introduced inductive bias and prior knowledge to extract local information for better feature extraction. Tokens-to-Token (T$2$T) ViT~\cite{yuan2021tokens} added a layer-wise T$2$T transformation and a deep-narrow backbone to overcome limitations of local structure modeling. Then, Han et al.~\cite{han2021transformer} proposed to model both patch- and pixel-level representations by transformer-in-transformer (TNT). Swin Transformer~\cite{liu2021swin} generated various patch scales by shifted windows for better representing highly changeable visual elements. Wu et al.~\cite{wu2021cvt} reintroduced convolutions into the ViT, namely convolutional vision transformer (CvT). Pyramid vision transformer (PVT)~\cite{wang2021pyramid} trained on dense partitions of the image to achieve high output resolution and used a progressive shrinking pyramid to reduce computations of large feature maps. Another Twins ViT framework~\cite{chu2021twins} was proposed, which introduced spatially separable self-attention (SSSA) to replace the less efficient global sub-sampled attention in PVT. All the aforementioned transformer structures were manually designed according to expert experience.

\textbf{One-shot NAS Method.} Differentiable architecture search (DARTS)~\cite{darts,yang2020ista,yang2021towards,huang2020explicitly} first formulated the NAS task in a differentiable manner based on the continuous relaxation. In contrast, single path one-shot (SPOS) framework \cite{guo2020single} adopted an explicit path sampler to construct a simplified supernet, such as uniform sampler \cite{guo2020single,su2021k}, greedy sampler \cite{you2020greedynas,huang2021greedynasv2} and Monte-Carlo tree sampler \cite{su2021prioritized}. Some work also attempted to investigate the channel dimension by direct searching \cite{su2021bcnet,locally} or pruning from pretrained models \cite{liu2017learning,tang2020reborn}. As for ViTs, AutoFormer~\cite{chen2021autoformer} first adopted the one-shot NAS framework for the ViT based architecture search.  BossNAS~\cite{li2021bossnas} implemented the search with an self-supervised training scheme and leveraged a hybrid CNN-transformer  search space for boosting the performance.

\begingroup
\setlength{\tabcolsep}{2pt}
\begin{table}[t]
\vspace{-1mm}
	\caption{Macro transformer space for the ViTAS of Twins-based architecture.``TBS'' indicates that layer type is searched from parametric operation and identity operation for depth search. ``Embedding'' represents the patch embedding layer.  ``$\text{Max}_\text{a}$'' and ``$\text{Max}_\text{m}$'' indicates the max dimension of attention layer (``$\text{Max}_\text{a}$'' also used for patch embedding layer) and mlp layer, respectively. ``Ratio'' means the reduction ratio from the ``Max Output Dim''. A larger ``Ratio'' indicates a larger dimension.}
	\label{search_space_table}
	\centering
	\vspace{-2mm}
    \resizebox{1\linewidth}{!}{
	\begin{tabular}{cccccccccc} 
	    \hline
		Number & OP & Type & Patch size / \#Heads & $\text{Max}_\text{a}$ & $\text{Max}_\text{m}$ & Ratio \\ 
		\hline
		$1$ & False & Embeding & 4 & $128$ & - & $\{i/10\}_{i=1}^{10}$\\ 
		\multirow{2}*{$4$} & \multirow{2}*{TBS} & Local & \multirow{2}*{$\{2, 4, 8, 16\}$} & \multirow{2}*{$480$} & \multirow{2}*{$512$} & \multirow{2}*{$\{i/10\}_{i=1}^{10}$} \\
		&  & Global & & & & \\ \hline
		$1$ & False & Embeding & 2 & $256$ & - & $\{i/10\}_{i=1}^{10}$\\
		\multirow{2}*{$4$} & \multirow{2}*{TBS} & Local & \multirow{2}*{$\{2, 4, 8, 16\}$} & \multirow{2}*{$960$} & \multirow{2}*{$1024$} & \multirow{2}*{$\{i/10\}_{i=1}^{10}$} \\
		&  & Global & & & & \\ \hline
		$1$ & False & Embeding & 2 & $512$ & - & $\{i/10\}_{i=1}^{10}$\\
		\multirow{2}*{$12$} & \multirow{2}*{TBS} & Local & \multirow{2}*{$\{2, 4, 8, 16\}$} & \multirow{2}*{$1920$} & \multirow{2}*{$2048$} & \multirow{2}*{$\{i/10\}_{i=1}^{10}$} \\
		&  & Global & & & & \\ \hline
		$1$ & False & Embeding & 2 & $1024$ & - & $\{i/10\}_{i=1}^{10}$\\
		\multirow{2}*{$6$} & \multirow{2}*{TBS} & Local & \multirow{2}*{$\{2, 4, 8, 16\}$} & \multirow{2}*{$3840$} & \multirow{2}*{$4096$} & \multirow{2}*{$\{i/10\}_{i=1}^{10}$} \\
		&  & Global & & & & \\
		\hline
	\end{tabular}}
	\vspace{-8mm}
\end{table}
\endgroup

\section{Revisiting One-shot NAS towards Transformer Space}

\textbf{One-shot NAS \& dimension search}. Towards the search of a decent architecture $\alpha \in \A$ from a huge transformer space $\A$ (\ie, transformer space), a weight sharing strategy is commonly leveraged to avoid exhausted path training from scratch. For a superformer $\N$ with weights $\W$, each path $\alpha$ inherits its weights from $\W$. The one-shot NAS is thus formulated as a two-stage optimization problem,~\ie, superformer training and then architecture searching. Base on the above settings, many researchers leveraged dimension search algorithms,~\eg, AutoSlim~\cite{autoslim} and BCNet~\cite{su2021bcnet}, to perform the search of the dimensions for fine grained architectures. We define $\C$ as the set of candidate dimensions for a certain operation, where $\c \in \C$ indicates the dimensions within $\alpha$. Thus, the optimization function is as
\begin{align}
    W_{\mathcal{A},\C}^*&=\argmin_{W_{\mathcal{A},\C}} loss_{\text{train}}\left(\mathcal{N}(\mathcal{A},\C,W_{\mathcal{A},\C})\right),\label{eq:trainsupernet}\\
    \balpha^*,\c^*&=\argmax_{(\balpha,\c)\in(\mathcal{A},\C)}Acc_{\text{val}}\left(\mathcal{N}(\alpha,\c,W_{\balpha,\c}^*)\right),\label{eq:searchoptimalpath}\\
    \text{s.t.}&\;\;\text{FLOPs}(\mathcal{N}(\alpha,\c,W_{\alpha,\c}^*))\le f,\nonumber
\end{align}
where $loss_{\text{train}}$ is training loss, $Acc_{\text{val}}$ is validation accuracy, $W_{\mathcal{A},\C}^*$ is a set of trained weights, $\alpha^*$ is the searched optimal architecture, and $f$ is resource budget. Following the one-shot framework, the superformer is trained by uniformly sampling different $(\alpha,\c)$ from $(\mathcal{A}, \C)$, and then we search the optimal architecture $(\alpha^*, \c^*)$ according to $W^*$. After these, the selected $\alpha^*$ will be retrained for evaluation.

\textbf{Towards Transformer Space.} To explore the possibility of the optimal ViT architecture in the arch-level, we incorporate all the essential elements in our transformer space, including \emph{head number}, \emph{patch size}, \emph{operation type}, \emph{output dimension of each layer}, and \emph{depth of the architectures}, as shown in Table~\ref{search_space_table}\footnote{In the superformer, $\text{Max}_\text{a}$ indicates the output of the first fully connected (FC) layer, which should be able to be divided by all ``ratios'' and ``Heads'', \ie, $\text{Max}_\text{a}|(Ratio \times Heads), \forall Ratio, Heads$. Therefore, we select least common multiple of ``ratios'' and ``Heads'' for $\text{Max}_\text{a}$} and Table~\ref{search_space_table2}\footnote{In the superformer, ``Max Dim'' indicates the output dimensions of both attention and MLP blocks.}. More details of transformer space is elaborated in the supplementary material. With the Twins-small based transformer space in Table~\ref{search_space_table} as an example, the size of transformer space amounts to $1.1 \times 10^{54}$ and the FLOPs (parameters) ranges from $0.02$G ($0.16$M) to $11.2$G ($86.1$M). Similarly in Table~\ref{search_space_table2}, with the DeiT-small transformer space, the size of space amounts to $5.4 \times 10^{34}$ and the FLOPs (parameters) ranges from $0.1$G ($0.5$M) to $20.0$G ($97.5$M).
\vspace{-1mm}

\begingroup
\setlength{\tabcolsep}{2pt}
\begin{table}[t]
\vspace{-1mm}
	\caption{Macro transformer space for the ViTAS of DeiT-based architecture.``TBS'' indicates that layer type is searched from vanilla ViT block or identity operation for depth search. ``Ratio'' means the reduction ratio from the ``Max Dim''. A larger ``Ratio'' means a larger dimension.}
	\label{search_space_table2}
	\centering
	\vspace{-2mm}
	\small{
	\begin{tabular}{cccccccccc} 
	    \hline
		Number & OP & Type & Patch size / \#Heads & Max Dim & Ratio \\ 
		\hline
		$1$ & False & Linear & $\{14, 16, 32\}$ & $384$ & $\{i/10\}_{i=1}^{10}$\\ 
		\multirow{2}*{$16$} & \multirow{2}*{TBS} & MHSA & $\{3, 6, 12, 16\}$ & $1440$ & $\{i/10\}_{i=1}^{10}$ \\
		&  & MLP & - & $1440$ & $\{i/10\}_{i=1}^{10}$ \\ 
		\hline
	\end{tabular}}
	\vspace{-4mm}
\end{table}
\endgroup

\section{Cyclic Channels for Token Embeddings} \label{cyclic_channels}

Previous work~\cite{autoslim} proposed the ordinal weight sharing paradigm, which is widely leveraged in many CNN and Transformer NAS papers~\cite{chen2021autoformer,wan2020fbnetv2,yan2021fp}. Concretely, as illustrated in Figure~\ref{fig:loop}(a), to search for a dimension $i$ at a layer with maximum of $l$ channels, the ordinal pattern assigns the left $i$ channels in the superformer to indicate the corresponding architecture as
\begin{equation}
    \a_A(i) = [1:i],~i \le l, \label{channel_assign_autoslim}
\end{equation}
where $\a_A(i)$ means the selected $i$ channels from the left (smaller-index) side.

However, this channel configuration imposes a strong constraint on the channels and leads to imbalanced training for each channel in the superformer. As in Figure~\ref{fig:loop}(a), with the ordinal pattern, channels that are close to the left side are used in both large and small dimension. Since different dimensions are uniformly sampled during searching, the training times $\C_A(i)$ of the $i$-th channel used in all dimensions with the ordinal pattern can be represented as
\begin{equation}
    \C_A(i) = l-i+1.\label{channel_usage_autoslim}
\end{equation}
Therefore, channels closer to the left side will gain more times of training, which induces evaluation bias among different channels and leads to sub-optimal searching results.


To remove the evaluation bias among channels of the superformer, we introduce a condition for constructing a mapping for channels:

\begin{theorem}[training fairness \cite{su2021bcnet}] \label{condition1} Each channel should obtain same training times for fairer training of superformer. 
\end{theorem}

With the aims of Condition~\ref{condition1} and keep the same computation cost as AutoSlim~\cite{autoslim} (\ie, ordinal pattern), we introduce indicator matrix $\beta$ with $\beta_{i,j} \in \{0, 1\}$ (one means using the channel) to represent whether channel $i$ being used in dimension $j$. Two conditions need to be satisfied: ($1$) for each row $\beta_i$, which is the training times of the channel $i$ in each dimension, the sum of it should be equal with that of all the other channels, and ($2$) for each column $\beta_j$, which demonstrates the training times of channels in dimension $j$, the sum of it should be the dimension of itself. Finally, the constraints of $\beta$ can be represented as follows
\vspace{-1mm}
\begin{align}
    \sum_j \beta_{i,j}&=(1 + l)/2,~\forall i,\label{weights}\\ 
    \sum_i \beta_{i,j}&=j,~\forall j.\label{widths}
\end{align}
\vspace{-1mm}
Infinite solutions can be solved under aforementioned constraints only. Here, the bilateral pattern in BCNet~\cite{su2021bcnet} is a special case of the aforementioned settings with double training times as $l+1$.


Although forcing the channels to be trained for same training times can boost the fairness, constructing a path only constrained by condition~\ref{condition1} cannot emerge the actual performance of the path due to the difference of training saturation between one-shot-based sampling and training from scratch~\cite{liu2018rethinking}. 
This means that in order to more precisely rank various paths, we need to mimic the process of the latter and balance the influence of each channels. 

\begin{theorem}[influence uniformity]\label{condition2}
Each channel needs to have the same sum of influence among all its related dimensions for training.
\end{theorem}

Concretely, we should carefully design the weight sharing mechanism based on \textbf{Condition~\ref{condition2}}. Here, we define $\psi_{i,j}$ to indicate the influence of channel $i$ in dimension $j$. For each channel pair, we have
\begin{equation}
    \left\{
        \begin{array}{rl}
            \psi_{i_1,j}=\psi_{i_2,j},&\forall i_1,i_2\\
            \psi_{i,j_1}\geq \psi_{i,j_2},&\forall j_1\leq j_2
        \end{array}
    \right..\label{influence}
\end{equation}
Considering that transformer architectures are mainly consist of full-connected (FC) layers. Specifically, for an FC layer with $j$ input channels $x_i,i=1,\cdots,j$ and a certain output channel $y=\sum_{i=1}^j y_i, y_i=w_i x_i$ with parameters $w_i,i=1,\cdots,j$, we can obtain the gradient of $w_i$ as 
\begin{equation}
    \bigtriangledown w_i= \frac{\partial loss_{\text{train}}}{\partial y}\cdot\frac{\partial y}{\partial y_i}\cdot\frac{\partial y_i}{\partial x_i}.
\end{equation}
Meanwhile, for a dimension sampled from the FC layer with one random input channel $x_i$, the gradient of $w_i$ here can be represented as
\begin{equation}
    (\bigtriangledown w_i)'= \frac{\partial loss_{\text{train}}}{\partial y'}\cdot\frac{\partial y'}{\partial y_i}\cdot\frac{\partial y_i}{\partial x_i},
\end{equation}
where $y'=y_i$. Therefore, in the former case, the influence $\psi_{i,j}$ of the $i$-th channel can be defined as the contribution of the channel to its gradient as
\begin{equation}
    \psi_{i,j}=\frac{(\bigtriangledown w_i)'}{\bigtriangledown w_i}=\frac{\frac{\partial y'}{\partial y_i}}{\frac{\partial y}{\partial y_i}},\label{eq:influence}
\end{equation}
assuming $\frac{\partial loss_{\text{train}}}{\partial y}=\frac{\partial loss_{\text{train}}}{\partial y'}$. We can also assume $y_i\approx y_{i'},i\neq i'$, as the distributions of $w_i$ or $x_i$ can be similar to each $i$, respectively, when randomly sampling the dimensions in each batch. In this case, we can obtain $y\approx j\times y_i$, and $\psi_{i,j} =\frac{1}{j}$ for Eq.~\eqref{influence} and~\eqref{eq:influence}. 

Note that channel $i$ may be shared~\wrt different dimensions. To keep all the channels being treated equally, any two channels $i_1$ and $i_2$ should have the same influence among all the dimensions,~\ie,
\begin{equation}
    \sum_j \beta_{i_1,j}\psi_{i_1,j} = \sum_j \beta_{i_2,j}\psi_{i_2,j},~\forall i_1,i_2.\label{eq_influence}
\end{equation}

\textbf{Optimization of cyclic mapping.} Combining Eq.~\eqref{weights} $\sim$ Eq.~\eqref{eq_influence}, we can obtain the specialized weight sharing paradigm for the cyclic superformer. In practice, since $\frac{1 + l}{2}$ may not be an integer, Eq.~\eqref{weights} may not be completely satisfied. Thus, for any two channels $i_1$ and $i_2$, we can relax the constraint in Eq.~\eqref{weights} by
\begin{equation}
    \left|\sum_j \beta_{i_1,j} - \sum_j \beta_{i_2,j} \right| \leq 1,~\forall i_1,i_2.\label{weights_release}
\end{equation}

To facilitate the search of the optimal weight sharing paradigm, we should make sure all the channels being fairly trained with almost the same influence among all the dimensions. Therefore, we can update Eq.~\eqref{eq_influence} to an objective as 
\begin{equation}
    \mathop{\min}_{\beta} \sum_{i_1,i_2} \left(\beta_{i_1,j}\psi_{i_1,j} - \beta_{i_2,j}\psi_{i_2,j}\right)^2,\label{eq_objective}
\end{equation}
The overall problem is thus a QCQP (quadratically constrained quadratic program), which can be efficiently solved by many off-the-shelf solvers~\cite{cvxpy, qcqp}. We have presented detailed experimental settings and simulations of $\beta$ and $\psi_{i,j}$ in supplementary materials.

\begin{figure}[!t]
    \centering
    \includegraphics[width=0.65\linewidth,height=0.65\linewidth]{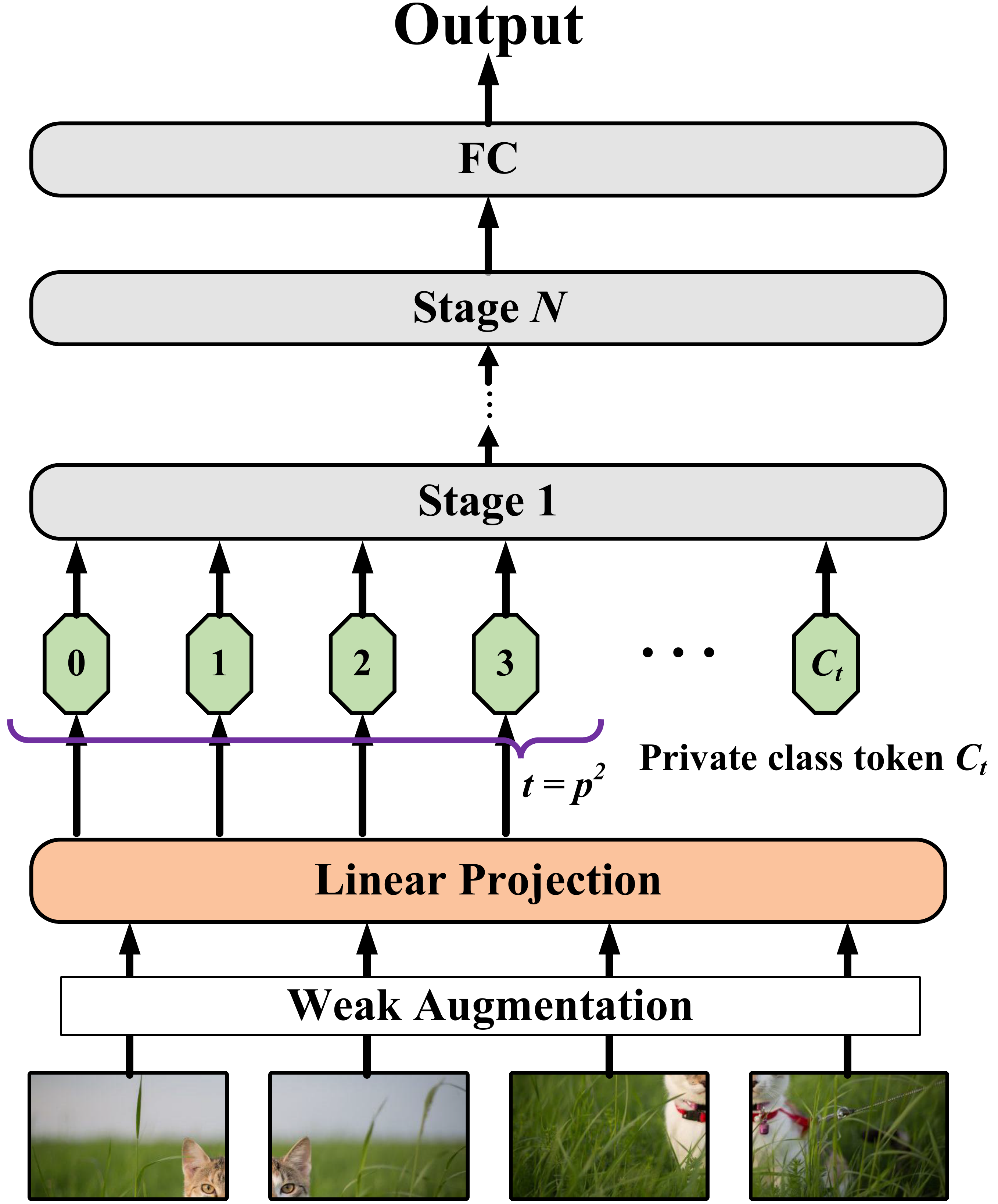}
    \caption{For transformer architectures with class token settings, with different patch sizes $p$, we assign one independent class tokens for each and obtain $p\times p$ patches and a private class token under one patch size setting.}\label{fig:class_token}
    \vspace{-4mm}
\end{figure}

\section{Further Stabilizing the Training of Superformer} \label{further}

\begin{figure}[!t]
    \centering
    \includegraphics[width=1\linewidth]{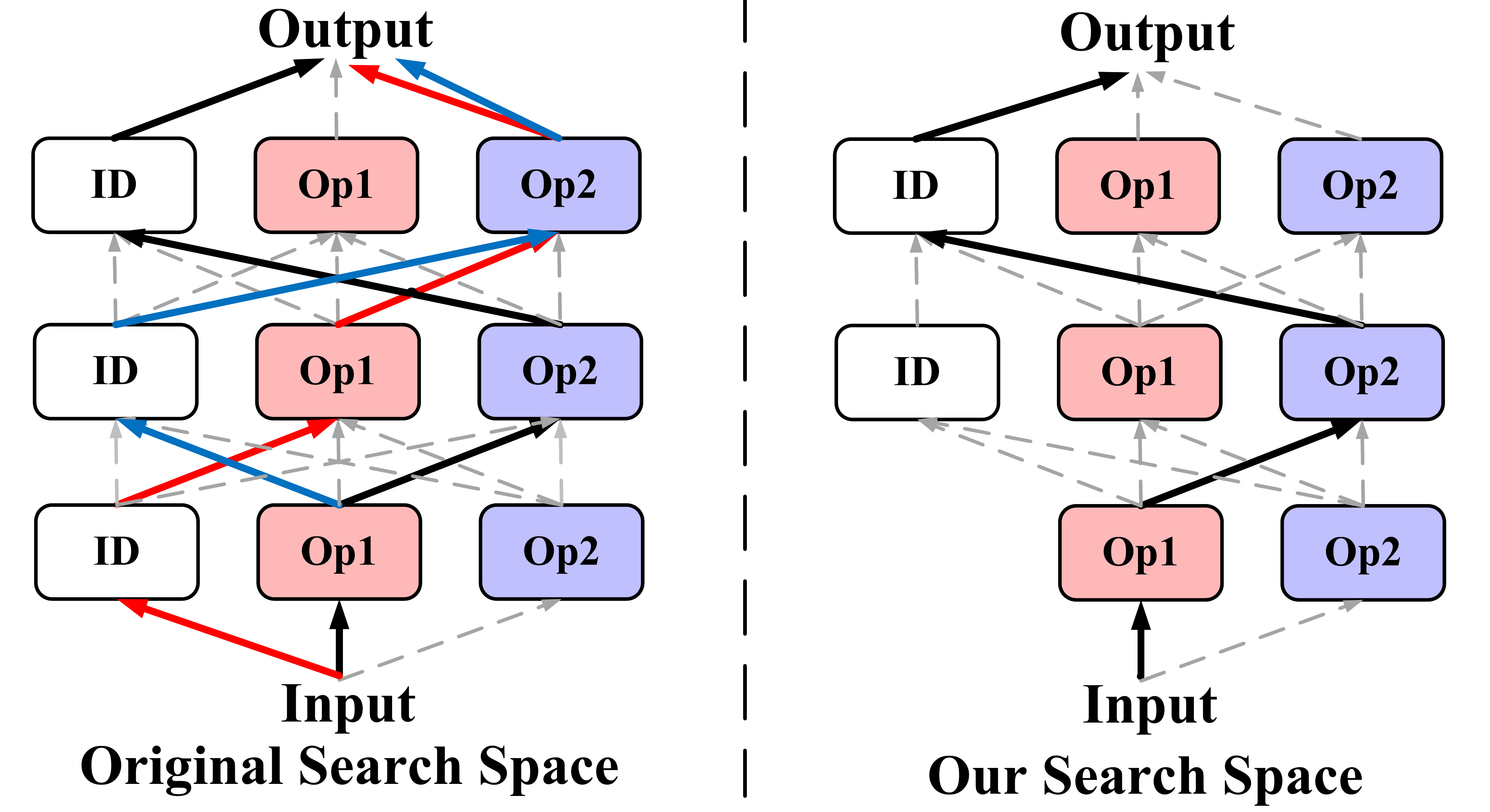}
    \caption{Comparison between currently mainstream ID search strategy (left) and ours ID oriented method (right). From left, a group of paths in the superformer,~\ie, {\color{red}\textbf{red}}, {\color{blue}\textbf{blue}} and \textbf{black} paths, corresponds to one same architecture, which may performs differently. This can hinder the search for decent architectures. While for our method, each architecture corresponds to only one path in the superformer, which reduces the redundancy in it and boosts the search performance.}\label{fig:id}
    \vspace{-4mm}
\end{figure}

Training the superformer is for fair estimation of each architecture's performance, which is essential for the next optimal architecture searching stage. Here, we argue that superformer requires an efficient and simple transformer space and training recipe for boosting the search. For the transformer space of ViTAS, we propose the identity shifting strategy to solve the many-to-one issue as in Figure \ref{fig:id}. Besides, for architectures with class tokens, \ie DeiT~\cite{touvron2020training}, we introduce private class token~\wrt~each patch size to cater for different paths. For the training recipe of ViTAS, we underline that weak augmentation $\&$ regularization rather  than complex and tricky ones \cite{touvron2020training,chu2021twins} can prevent the search from unsteady.

\textbf{Identity shifting.} Given a pre-defined NAS transformer space, identity (ID) operation serves as the significant part and has a large effect on the searched results for three reasons: $1$) it defines the depth of the searched architecture, $2$) compared to other operations, the non-parametric ID is much more different with other parametric operations, which will involve in a higher variance on paths (as in Figure~\ref{fig:id}(a)), and $3$) stacking manner and intrinsic structure of a transformer architecture within each stage lead to complicatedly many-to-one correspondence between architectures in the superformer and transformer space (see Figure~\ref{fig:id}(a)). These introduce a huge ambiguous for NAS. Here, we propose to search the operations with identity shifting strategy, as depicted in Figure~\ref{fig:id}(b). In each stage, we remove the ambiguous between transformer space and superformer by sampling the number of ID and arrange them at the deeper layers of the stage in order to remove the redundancy in superformer. Typically, with three operations (including ID) and twelve searched layers, the transformer space of operations can be reduced from $3^{12}$ to $2^{13}-1$.

\noindent\textbf{Private class token.} Notably, pure vision transformer architectures,~\eg, DeiT~\cite{touvron2020training}, usually introduce a trainable vector named class token for the output classification. The class token is appended to the patch tokens before the first layer and then go through the transformer blocks for the prediction task. These class tokens often take a small size, which changes with the pre-defined patch size $P$ (\ie, $\frac{H \times W}{P \times P}$), and performs significant in performance. Towards these attributes, we propose to privatize the class token for each $P$. As shown in Figure~\ref{fig:class_token}, for different patch sizes, we assign private ones for each. In this way, the affect between class tokens can be avoided with only negligible computation cost or memory cost introduced.

\noindent\textbf{Weak augmentation \& regularization.} We explore the superformer training strategy of the ViTAS, including data augmentation and regularization. We conducted the evaluations with Twins-based architecture and $1.4$G FLOPs budget on ImageNet-1k dataset. Compared with one single ViT, the superformer training is much more difficult to converge, which needs a simply yet effective training strategy.

From Table~\ref{MCT_NAS_analysis}, as group $0$, the superformer performs badly with the default training strategy~\cite{touvron2020training,chu2021twins}. To facility the search, we first removed the stochastic depth, since our identity search performs the similar effect in the superformer training. Then, we gradually dropped other data augmentation and/or regularization and find that a weak augmentation can largely promote retraining accuracy with searching a better architecture. Table~\ref{Training_Recipe} presents the ViTAS training recipe for our experiments. 

\section{Experimental Results}\label{sec:results}

\begin{table}[t]
	\caption{Searched Twins-based ViT architectures~\wrt~different FLOPs and GPU throughput on ImageNet-$1$k. We abbreviate the name of tiny, short, base, and large for T, S, B, and L, respectively. $\star$ indicates that the re-implementation results of important baseline methods with our recipe. Our results are highlighted in bold.} 
	\label{tbl:sota_twins}
	\centering
	\small
	\vspace{-2mm}
	\setlength\tabcolsep{1.5pt}
	\begin{tabular}{r||ccc||ccccc} 
	    \hline
		Method & \tabincell{c}{FLOPs\\ (G)} & \tabincell{c}{Throughput\\ (image/s)} &\tabincell{c}{Params \\ (M)} & \tabincell{c}{Top-$1$\\ (\%)}&\tabincell{c}{Top-$5$\\ (\%)}\\ \hline
		\hline
		ResNet-$18$ \cite{dr} & $1.8$ & $4458.4$ & $12$ & $69.8$ & $89.1$ \\ 
		DeiT-T$^{\star}$ \cite{touvron2020training} & $1.3$ & $2728.5$ & $5$ & $72.3$ & $91.4$   \\
        Twins-T$^{\star}$ \cite{chu2021twins} & $1.4$ & $1580.7$ & $11.5$ & $77.8$ & $94.1$ \\
		AutoFormer-T$^{\star}$ \cite{chen2021autoformer} & $1.3$ & $3055.4$ & $5.7$ & $74.7$ & $91.9$ \\
		ViTAS-Twins-T & $1.4$ & $1686.3$ & $13.8$ & $\textbf{79.4}$ & $\textbf{94.8}$ \\  \hline
		ResNet-$50$ \cite{dr} & $4.1$ & $1226.1$ & $25$ &  $76.2$ & $91.4$ &  \\
		T2T-ViT$_t$-14 \cite{yuan2021tokens} & $5.2$ & - & $22.0$ & $80.7$ & - \\
		DeiT-S$^{\star}$ \cite{touvron2020training} & $4.6$ & $437.0$ & $22.1$ & $79.9$ & $95.0$ \\
		TNT-S \cite{han2021transformer} & $5.2$ & - & $23.8$ & $81.3$ & - \\
		PVT-S \cite{wang2021pyramid} & $3.8$ & $820$ & $24.5$ & $79.8$ & - \\
		Twins-SVT-S$^{\star}$ \cite{chu2021twins} & $2.9$ & $1059$ & $24$ & $81.6$ & $95.9$ \\
		CPVT-S-GAP \cite{chu2021conditional} & $4.6$ & $817$ & $23$ & $81.5$ & - \\
 		AutoFormer-S$^{\star}$ \cite{chen2021autoformer} & $5.1$ & $1231.7$ & $22.9$ & $79.8$ & $95.0$ \\ 
 		BossNet-T0 \cite{li2021bossnas} & $5.7$ & - & - & $81.6$ & - \\
		Twins-PCPVT-S$^{\star}$ \cite{chu2021twins} & $3.8$ & $815$ & $24.1$ & $81.2$ & $95.6$ \\
		Swin-T$^{\star}$ \cite{liu2021swin} & $4.5$ & $766$ & $29$ & $81.2$ & $95.5$ \\
		ViTAS-Twins-S & $3.0$ & $958.6$ & $30.5$ & $\textbf{82.0}$ & $\textbf{95.7}$ \\ \hline
		T2T-ViT$_t$-19 \cite{yuan2021tokens} & $8.9$ & - & $39.2$ & $81.4$ & - \\
		BoTNet-S1-59 \cite{srinivas2021bottleneck} & $7.3$ & - & $33.5$ & $81.7$ & - \\
		BossNet-T1 \cite{li2021bossnas} & $7.9$ & - & - & $82.2$ & - \\
		Twins-PCPVT-B \cite{chu2021twins} & $6.7$ & $525$ & $43.8$ & $82.7$ & - \\
		Swin-S$^{\star}$ \cite{liu2021swin} & $8.7$ & $444$ & $50$ & $83.0$ & $96.2$ \\
		Twins-SVT-B$^{\star}$ \cite{chu2021twins} & $8.6$ & $469$ & $56$ & $83.2$ & $96.3$\\
		ViTAS-Twins-B & $8.8$ & $362.7$ & $66.0$ & $\textbf{83.5}$ & $\textbf{96.5}$ \\  \hline
		DeiT-B$^{\star}$ \cite{touvron2020training} & $17.6$ & $292$ & $86.6$ & $81.8$ & $95.7$ \\
		TNT-B \cite{han2021transformer} & $14.1$ & - & $66$ & $82.8$ & - \\
		CrossViT-B \cite{chen2021crossvit} & $21.2$ & - & $104.7$ & $82.2$ & - \\
		ViTAS-Twins-L & $16.1$ & $260.7$ & $124.8$ & $\textbf{84.0}$ & $\textbf{96.9}$ \\  \hline
	    \hline
	\end{tabular}
	\vspace{-3mm}
\end{table}

\begingroup
\setlength{\tabcolsep}{2pt}
\begin{table}[t]
	\caption{Ablation studies of training recipe of superformer of the ViTAS. ``\Checkmark''/``\XSolidBrush'' indicates that we used/not used the corresponding method. We implemented the search on the ImageNet-$1$k with $1.4$G budget of Twins. We report the top-$1$ accuracy of the best architectures in both ViTAS (\ie, searching) and retraining. ``RD'': Rand-Augment. ``MP'': Mixup. ``CM'': CutMix. ``CJ'': Color Jitter. ``Era'': Erasing. ``SD'': Stoch Depth. ``RA'': Repeated Augmentation. ``WD'': Weight Decay. Best results are highlighted in bold.}
	\label{MCT_NAS_analysis}
	\centering
	\resizebox{1\linewidth}{!}{
	\begin{tabular}{c|cccc|cccc|c|c} \hline
		&\multicolumn{4}{c}{Data augmentation} & \multicolumn{4}{|c|}{Regularization} & ViTAS & Retraining\\ \cline{1-9}
		\#&RD & MP &  CM & CJ & Era & SD & RA & WD & Acc(\%)& Acc(\%) \\ \hline
		$0$& \Checkmark & \Checkmark& \Checkmark& \Checkmark& \Checkmark& \Checkmark&\Checkmark & \Checkmark &$59.4\%$ & $77.9\%$\\ 
		$1$ & \lightgray{\Checkmark}& \lightgray{\Checkmark}& \lightgray{\Checkmark}& \lightgray{\Checkmark}& \lightgray{\Checkmark}& \XSolidBrush & \lightgray{\Checkmark}& \lightgray{\Checkmark}&$61.2\%$ & $78.0\%$\\ 
		$2$& \lightgray{\Checkmark}& \lightgray{\Checkmark}& \lightgray{\Checkmark}& \lightgray{\Checkmark}& \lightgray{\Checkmark}& \lightgray{\XSolidBrush} & \XSolidBrush &\lightgray{\Checkmark}&$61.5\%$ & $78.2\%$\\ 
	    $3$& \lightgray{\Checkmark}& \XSolidBrush & \XSolidBrush & \lightgray{\Checkmark}&\lightgray{\Checkmark}& \lightgray{\XSolidBrush} & \lightgray{\XSolidBrush} &\lightgray{\Checkmark}&$62.3\%$ & $78.6\%$\\  
		$4$& \XSolidBrush & \lightgray{\XSolidBrush} & \lightgray{\XSolidBrush} & \lightgray{\Checkmark}&\lightgray{\Checkmark}& \lightgray{\XSolidBrush} & \lightgray{\XSolidBrush} &\lightgray{\Checkmark}&$64.9\%$ & $78.8\%$\\  
		$7$& \lightgray{\XSolidBrush} & \lightgray{\XSolidBrush} & \lightgray{\XSolidBrush} & \XSolidBrush &\lightgray{\Checkmark}& \lightgray{\XSolidBrush} & \lightgray{\XSolidBrush} &\lightgray{\Checkmark}&$65.6\%$ & $78.9\%$\\
		$8$& \lightgray{\XSolidBrush} & \lightgray{\XSolidBrush} & \lightgray{\XSolidBrush} & \lightgray{\XSolidBrush} & \XSolidBrush & \lightgray{\XSolidBrush} & \lightgray{\XSolidBrush} &\lightgray{\Checkmark}&$66.1\%$ & $79.1\%$\\
		$9$& \lightgray{\XSolidBrush} & \lightgray{\XSolidBrush} & \lightgray{\XSolidBrush} & \lightgray{\XSolidBrush} & \lightgray{\XSolidBrush} & \lightgray{\XSolidBrush} & \lightgray{\XSolidBrush} & \XSolidBrush &$\boldsymbol{67.7}\%$ & $\boldsymbol{79.4\%}$\\
		\hline
	\end{tabular}}
	\vspace{-2mm}
\end{table}
\endgroup

\begingroup
\setlength{\tabcolsep}{2pt}
\begin{table}[t]
	\caption{Training recipe of the ViTAS with parameter settings. BS: batch size, LR: learning rate, WD: weight decay. We will conduct the ViTAS according to the following recipe in experiments.}
	\label{Training_Recipe}
	\centering
	\begin{tabular}{cccccccc} \hline
		Epochs & BS & Optimizer & LR & LR decay & Warmup\\ 
		\hline
		$300$ & $1024$ & AdamW & $0.001$ & cosine & $5$ \\ 
		\hline
	\end{tabular}
	\vspace{-2mm}
\end{table}
\endgroup
We perform the ViTAS on the challenging ImageNet-$1$k dataset~\cite{Imagenet} for image classification, and COCO$2017$~\cite{COCO} and ADE$20$k~\cite{zhou2017scene} for object detection, instance segmentation, and semantic segmentation. To promote the search, we randomly sample $50$K images from the training set as the local validation set and the rest images are leveraged for training. All experiments are implemented with PyTorch~\cite{pytorch} and trained on NVIDIA Tesla V$100$ GPUs. Please find detailed experimental settings in the supplementary material.

\begin{table}[t]
	\caption{Searched ViT architectures that do not involve inductive bias~\wrt~different FLOPs and GPU throughput on ImageNet-$1$k. $\star$ indicates that the re-implementation results of important baseline methods with our recipe. Our results are highlighted in bold.} 
	\label{tbl:deit}
	\centering
	\setlength\tabcolsep{1.5pt}
	\small
	\begin{tabular}{r||ccc||ccccc} 
	    \hline
		Method & \tabincell{c}{FLOPs\\ (G)} & \tabincell{c}{Throughput\\ (image/s)} &\tabincell{c}{Params \\ (M)} & \tabincell{c}{Top-$1$\\ (\%)}&\tabincell{c}{Top-$5$\\ (\%)}\\ 
		\hline
		ResNet-$18$ \cite{dr} & $1.8$ & $4458.4$ & $12$ & $69.8$ & $89.1$ \\ 
		DeiT-T$^{\star}$ \cite{touvron2020training} & $1.3$ & $2728.5$ & $5$ & $72.3$ & $91.4$   \\
		AutoFormer-T$^{\star}$ \cite{chen2021autoformer} & $1.3$ & $2955.4$ & $5.7$ & $74.7$ & $91.9$ \\
		ViTAS-DeiT-A & $1.4$ & $2831.1$ & $6.6$ & $\textbf{75.6}$ & $\textbf{92.5}$ \\ \hline
 		ResNet-$50$ & $4.1$ \cite{dr} & $1226.1$ & $25$ &  $76.2$ & $91.4$ &  \\
  		DeiT-S$^{\star}$ \cite{touvron2020training} & $4.6$ & $940.4$ & $22$ & $79.9$ & $95.0$ &  \\
 		AutoFormer-S$^{\star}$ \cite{chen2021autoformer} & $5.1$ & $1231.7$ & $22.9$ & $79.8$ & $95.0$ \\
		ViTAS-DeiT-B & $4.9$ & $1189.4$ & $2.3$ & $\textbf{80.2}$ & $\textbf{95.1}$ \\
	    \hline
	\end{tabular}
	\vspace{-2mm}
\end{table}

\subsection{Efficient search of ViTAS on ImageNet-$1$k} \label{Imagenet}

\begin{table*}[t]
	\caption{Object detection and instance segmentation performance with searched backbones on the COCO$2017$ dataset with Mask R-CNN framework and RatinaNet framework. We followed the same training and evaluation setting as~\cite{chu2021twins}. ``FLOPs'' and ``Param'' are in giga and million, respectively.
	$\star$ indicates the re-implementation results of important baseline methods with our recipe. Our results are highlighted in bold.} 
	\label{tbl:maskrcnn}
	\centering
	\setlength\tabcolsep{2pt}
	\small
	\vspace{-3mm}
	\begin{tabular}{r||cc||cccccc||cc||cccccc} 
	    \hline
	    \multirow{2}*{Backbone} & \multicolumn{8}{c||}{Mask R-CNN 1$\times$ \cite{he2017mask}} & \multicolumn{8}{c}{RetinaNet 1$\times$ \cite{lin2017focal}} \\\cline{2-17}
	    & FLOPs & Param & $\text{AP}^{\text{b}}$ & AP$_{50}^{\text{b}}$ & AP$_{75}^{\text{b}}$ & AP$^{\text{m}}$ & AP$_{50}^{\text{m}}$ & AP$_{75}^{\text{m}}$ & FLOPs & Param & $\text{AP}^{\text{b}}$ & AP$_{50}^{\text{b}}$ & AP$_{75}^{\text{b}}$ & AP$_{S}$ & AP$_{M}$ & AP$_{L}$ \\ \hline
        ResNet50 \cite{dr} & $174$ & $44.2$ & $38.0$ & $58.6$ & $41.4$ & $34.4$ & $55.1$ & $35.7$ & 111 & 37.7 & 36.3 & 55.3 & 38.6 & 19.3 & 40.0 & 48.8 \\
        PVT-Small \cite{wang2021pyramid} & $178$ & $44.1$ & $40.4$ & $62.9$ & $43.8$ & $37.8$ & $60.1$ & $40.3$ & 118 & 34.2 & 40.4 & 61.3 & 43.0 & 25.0 & 42.9 & 55.7 \\
        Twins-PCPVT-S \cite{chu2021twins} & $178$ & $44.3$ & $42.9$ & $65.8$ & $47.1$ & $40.0$ & $62.7$ & $42.9$ & $118$ & $34.4$ & $43.0$ & $64.1$ & $46.0$ & $27.5$ & $46.3$ & $57.3$ \\
        Swin-T \cite{liu2021swin} & $177$ & $47.8$ & $42.2$ & $64.6$ & $46.2$ & $39.1$ & $61.6$ & $42.0$ & $118$ & $38.5$ & $41.5$ & $62.1$ & $44.2$ & $25.1$ & $44.9$ & $55.5$ \\
        Twins-SVT-S$^{\star}$ \cite{chu2021twins} & $164$ & $44.0$ & $43.5$ & $66.0$ & $47.8$ & $40.1$ & $62.9$ & $43.1$ & $104$ & $34.3$ & $42.2$ & $63.3$ & $44.9$ & $26.4$ & $45.6$ & $57.0$ \\
        ViTAS-Twins-S & 168 & 44.2 & $\textbf{45.9}$ & $\textbf{67.8}$ & $\textbf{50.3}$ & $\textbf{41.5}$ & $\textbf{64.7}$ & $\textbf{45.0}$ & $108$ & $41.3$ & $\textbf{44.4}$ & $\textbf{65.3}$ & $\textbf{47.6}$ & $\textbf{27.5}$ & $\textbf{48.3}$ & $\textbf{60.0}$ \\ \hline
        ResNet101 \cite{dr} & $210$ & $63.2$ & $40.4$ & $61.1$ & $44.2$ & $36.4$ & $57.7$ & $38.8$ & $149$ & $56.7$ & $38.5$ & $57.8$ & $41.2$ & $21.4$ & $42.6$ & $51.1$ \\
        ResNeXt101 \cite{xie2017aggregated} & $212$ & $62.8$ & $41.9$ & $62.5$ & $45.9$ & $37.5$ & $59.4$ & $40.2$ & $151$ & $56.4$ & $39.9$ & $59.6$ & $42.7$ & $22.3$ & $44.2$ & $52.5$ \\
        PVT-Medium \cite{chu2021twins} & $211$ & $63.9$ & $42.0$ & $64.4$ & $45.6$ & $39.0$ & $61.6$ & $42.1$ & $151$ & $53.9$ & $41.9$ & $63.1$ & $44.3$ & $25.0$ & $44.9$ & $57.6$ \\
        Twins-PCPVT-B \cite{chu2021twins} & $211$ & $64.0$ & $44.6$ & $66.7$ & $48.9$ & $40.9$ & $63.8$ & $44.2$ & $151$ & $54.1$ & $44.3$ & $65.6$ & $47.3$ & $27.9$ & $47.9$ & $59.6$ \\
        Swin-S \cite{liu2021swin} & $222$ & $69.1$ & $44.8$ & $66.6$ & $48.9$ & $40.9$ & $63.4$ & $44.2$ & $162$ & $59.8$ & $44.5$ & $65.7$ & $47.5$ & $27.4$ & $48.0$ & $59.9$ \\
        Twins-SVT-B$^{\star}$ \cite{chu2021twins} & $224$ & $76.3$ & $45.5$ & $67.4$ & $50.0$ & $41.4$ & $64.5$ & $44.5$ & $163$ & $67.0$ & $44.4$ & $65.6$ & $47.4$ & $28.5$ & $47.9$ & $59.5$ \\
        ViTAS-Twins-B & 227 & 85.4 & $\textbf{47.6}$ & $\textbf{69.2}$ & $\textbf{52.2}$ & $\textbf{42.9}$ & $\textbf{66.3}$ & $\textbf{46.5}$ & $167$ & $76.2$ & $\textbf{46.0}$ & $\textbf{66.7}$ & $\textbf{49.6}$ & $\textbf{29.1}$ & $\textbf{50.2}$ & $\textbf{62.0}$ \\ \hline
        Twins-SVT-L$^{\star}$ \cite{chu2021twins} & 292 & 119.7 & 45.9 & 67.9 & 49.9 & 41.6 & 65.0 & 45.0 & 232 & 110.9 & 45.2 & 66.6 & 48.4 & 29.0 & 48.6 & 60.9 \\  
        ViTAS-Twins-L & 301 & 144.1 & $\textbf{48.2}$ & $\textbf{69.9}$ & $\textbf{52.9}$ & $\textbf{43.3}$ & $\textbf{66.9}$ & $\textbf{46.7}$ & 246 & 135.5 & \textbf{47.0} & \textbf{67.8} & \textbf{50.3} & \textbf{29.6} & \textbf{50.9} & \textbf{62.4} \\ \hline
	    \hline
	\end{tabular}
	\vspace{-1mm}
\end{table*}

In Table~\ref{tbl:sota_twins}, we compare our results with recent precedent ViT architectures. To evaluate our methods with other existing algorithms, we based on the Twins transformer space\footnote{We constructed the ViTAS-Twins-T transformer space from Twins-S similar to Table~\ref{search_space_table}, and the Twins-T was uniformly scaled from Twins-S.} and present the search results with both FLOPs and GPU throughput. With different FLOPs budgets, The search ones (ViTAS-Twins-T/S/B/L) can outperform the referred transformers. For example, ViTAS-Twins-T can improve Top-$1$ accuracy by $1.6\%$, compared with Twins-T. For larger architectures, the search ones can moderately surpass all the corresponding referred ones as well, respectively.

In addition, we also searched the optimal architectures based on pure ViT and DeiT space, shown in Table~\ref{tbl:deit}. With only $1.4$G FLOPs and similar GPU throughput, our searched ViTAS-DeiT-A model achieves $75.6\%$ on Top-1 accuracy and is $3.4\%$ superior than DeiT-T, which indicates the effectiveness of our proposed ViTAS method. Furthermore, with $4.9$G FLOPs budget, our ViTAS-DeiT-B model also achieves superior performance of $80.2\%$ on Top-$1$ accuracy with $0.3$\% surpassing the DeiT-S.

\begin{table}[t]
	\caption{Performance comparisons with searched backbones on ADE$20$K validation dataset. Architectures were implemented with the same training recipe as~\cite{chu2021twins}. All backbones were pretrained on ImageNet-$1$k, except for SETR, which was pretrained on ImageNet-$21$k dataset. $\star$ indicates the re-implementation results of important baseline methods with our recipe. Our results are highlighted in bold.} 
	\label{tbl:ADE$20$k}
	\centering
	\setlength\tabcolsep{2pt}
	\small
	\vspace{-2mm}
	\resizebox{1\linewidth}{!}{
	\begin{tabular}{r||ccc||ccc} 
	    \hline
	    \multirow{3}*{Backbone} & \multicolumn{3}{c||}{Semantic FPN 80k \cite{chu2021twins}} & \multicolumn{3}{c}{Upernet 160k \cite{liu2021swin}} \\
	    & FLOPs & Param & mIoU & FLOPs & Param & mIoU \\
	    & (G) & (M) & (\%) & (G) & (M) & (\%) \\ \hline
        ResNet50 \cite{dr} & $45$ & $28.5$ & $36.7$ & - & - & - \\
        PVT-Small \cite{wang2021pyramid} & $40$ & $28.2$ & $39.8$ &  - & - & - \\
        Twins-PCPVT-S \cite{chu2021twins} & $40$ & $28.4$ & $44.3$ & $234$ & $54.6$ & $46.2$ \\
        Swin-T \cite{liu2021swin} & $46$ & $31.9$ & $41.5$ & $237$ & $59.9$ & $44.5$ \\
        Twins-SVT-S$^{\star}$ \cite{chu2021twins} & $37$ & $28.3$ & $43.6$ & $228$ & $54.4$ & $45.9$ \\
        ViTAS-Twins-S & 38 & 35.1 & $\textbf{46.6}$ & 229 & 61.7 & $\textbf{47.9}$ \\
        \hline
        ResNet101 \cite{dr} & $66$ & $47.5$ & $38.8$ & - & - & - \\
        PVT-Medium \cite{wang2021pyramid} & $55$ & $48.0$ & $41.6$ & - & - & - \\
        Twins-PCPVT-B \cite{chu2021twins} & $55$ & $48.1$ & $44.9$ & $250$ & $74.3$ & $47.1$ \\
        Swin-S \cite{liu2021swin} & $70$ & $53.2$ & $45.2$ & $261$ & $81.3$ & $47.6$ \\
        Twin-SVT-B$^{\star}$ \cite{chu2021twins} & $67$ & $60.4$ & $45.5$ & $261$ & $88.5$ & $47.7$ \\ 
        ViTAS-Twins-B & 67 & 69.6 & $\textbf{49.5}$ & 261 & 97.7 & $\textbf{50.2}$ \\
        \hline
        ResNetXt101 \cite{xie2017aggregated} & - & $86.4$ & $40.2$ & - & - & - \\
        PVT-Large \cite{chu2021twins} & $71$ & $65.1$ & $42.1$ & - & - & - \\
        Twins-PCPVT-L \cite{chu2021twins} & $71$ & $65.3$ & $46.4$ & $269$ & $91.5$ & $48.6$ \\
        Swin-B \cite{liu2021swin} & $107$ & $91.2$ & $46.0$ & $299$ & $121$ & $48.1$ \\
        Twins-SVT-L$^{\star}$ \cite{chu2021twins} & $102$ & $103.7$ & $46.9$ & $297$ &  $133$ & $48.8$ \\ 
        ViTAS-Twins-L & 108 & 128.2 & $\textbf{50.4}$ & 303 & 158.7 & $\textbf{51.3}$ \\ \hline
        Backnone & \multicolumn{3}{c||}{PUP (SETR \cite{zheng2021rethinking})} & \multicolumn{3}{c}{MLA (SETR \cite{zheng2021rethinking})} \\ \hline
        T-Large (SETR) \cite{zheng2021rethinking} & - & $310$ & $50.1$ & - & $308$ & $48.6$ \\
	    \hline
	\end{tabular}} 
	\vspace{-3mm}
\end{table}

\subsection{Transferability of ViTAS with Semantic Segmentation on ADE$20$K} \label{transfer_ADE$20$k}

In addition to search the optimal ViT architectures on the ImageNet-$1$k, we evaluated the generalization ability of the ViTAS by transferring the searched architectures to other tasks. With the same recipe as Twins~\cite{chu2021twins}, we fintuned on ADE$20$k~\cite{zhou2017scene} by using our ImageNet-pretrained models as backbones for semantic segmentation, shown in Table~\ref{tbl:ADE$20$k}. Under different FLOPs budgets, our models can obtain significant performance improvement as $2\%\sim4\%$ on mIoU, compared with corresponding referred methods. For example, with semantic FPN~\cite{chu2021twins} method as baseline, ViTAS-Twins-B surpasses the second best one, Twins-SVT-B, by more than $4\%$ on mIoU.

\subsection{Transferability to Object Detection and Instance Segmentation} \label{transfer_COCO$2017$}

With the same recipe as Twins~\cite{chu2021twins}, we undertook both object detection and instance segmentation on COCO$2017$~\cite{lin2014microsoft} by using our ImageNet-pretrained models as backbones, respectively. In Table~\ref{tbl:maskrcnn}, with mask R-CNN and RetinaNet as baseline, we achieve state-of-the-art performance with remarkably improvement on each AP metrics. The aforementioned experimental results in both of the tasks can demonstrate the effectiveness of ViTAS.

\begin{figure}[!t]
    \centering
    
	\begin{subfigure}[t]{.48\linewidth}
        \centering
        \includegraphics[width=1\linewidth]{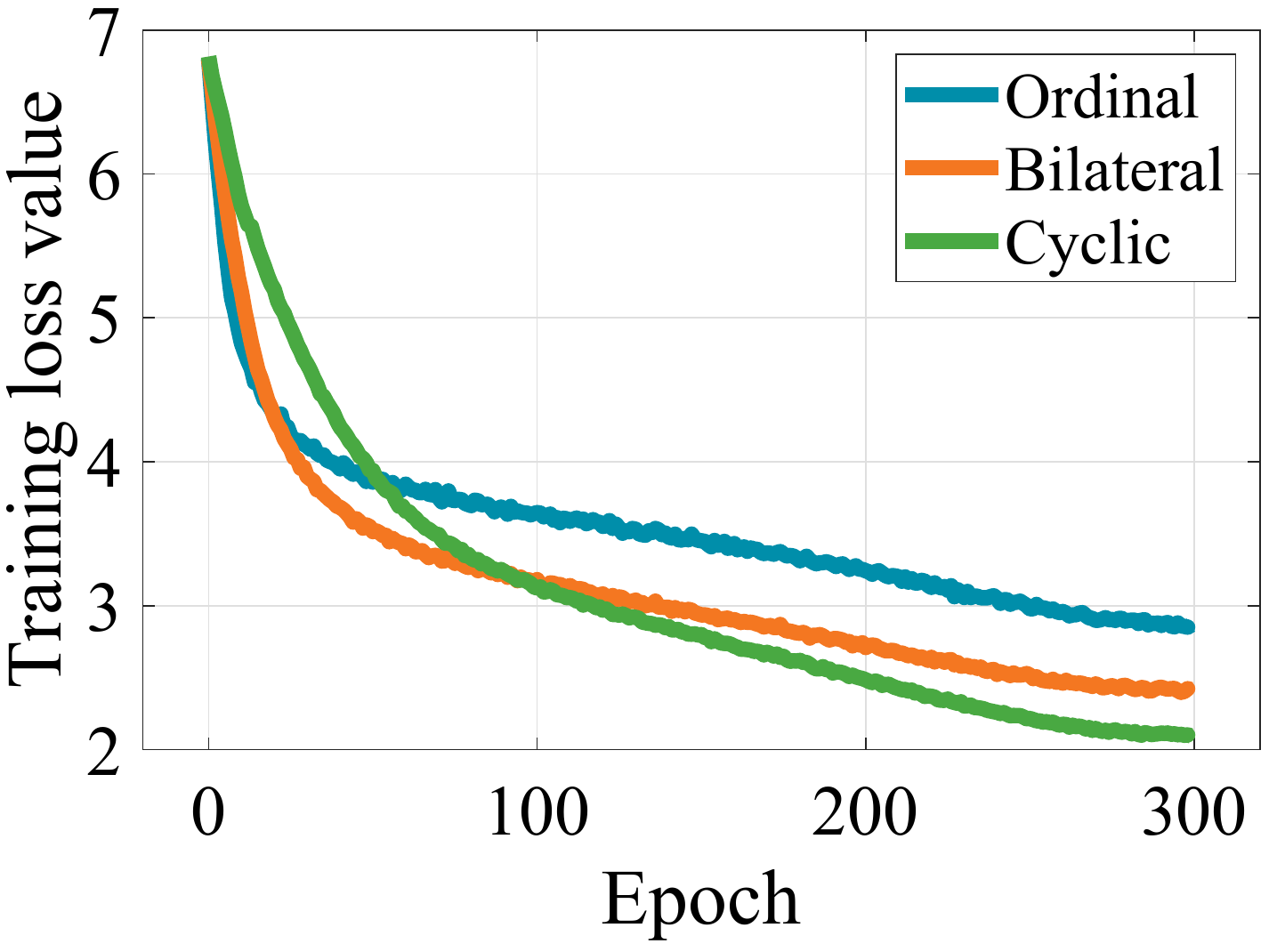}
        \subcaption{Weight sharing paradigm.}
    \end{subfigure}
	\begin{subfigure}[t]{.48\linewidth}
        \centering
        \includegraphics[width=1\linewidth]{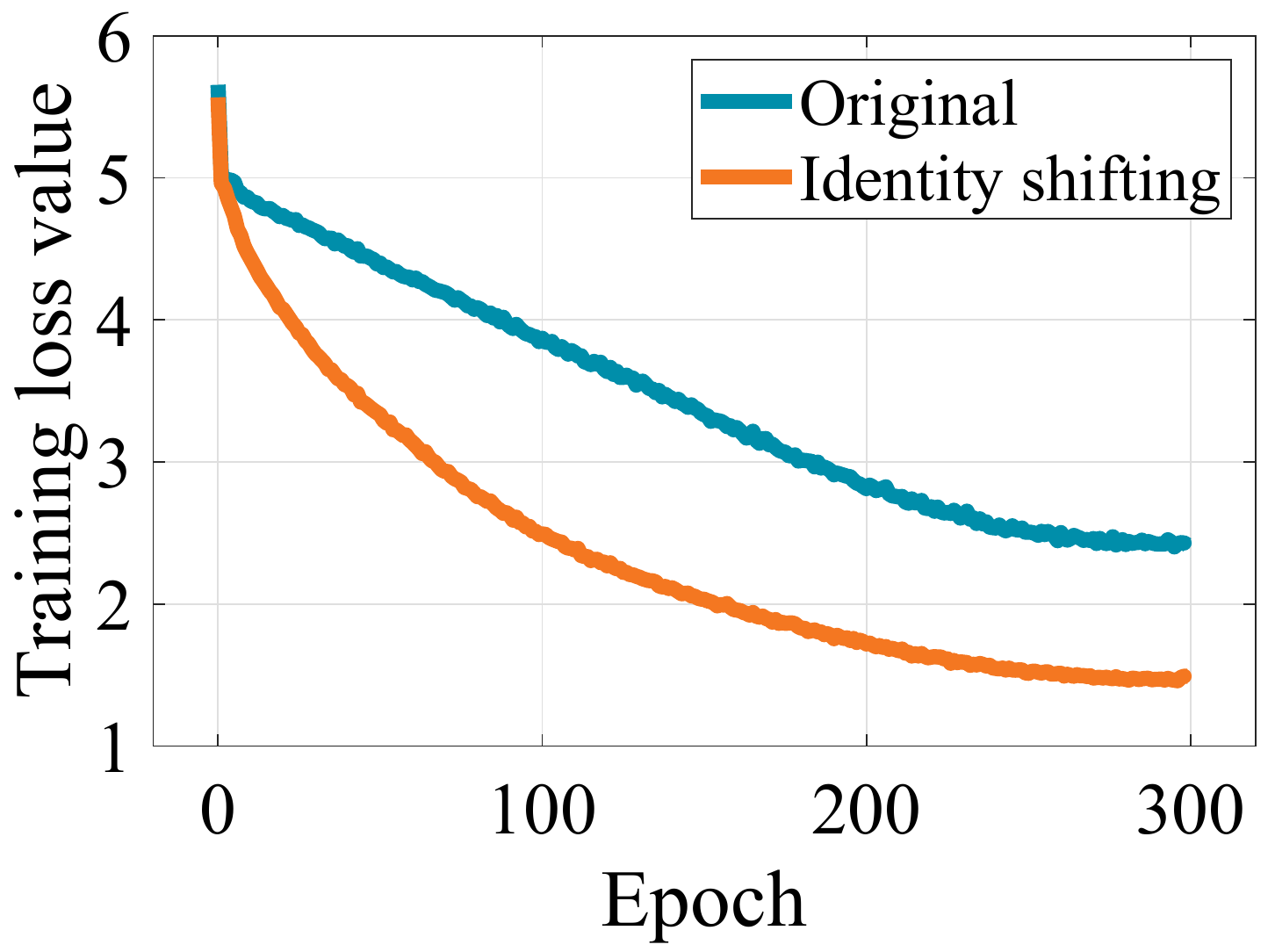}
        \subcaption{Identity shifting.}
    \end{subfigure}
    \caption{Comparisons of superformer training losses. (a) Superformer training loss~\wrt~ordinal, bilateral, and cyclic weight sharing mechanisms on ImageNet-$1$k dataset. (b) Superformer training loss~\wrt~identity shifting and original setting on ImageNet-$100$.}\label{fig:training_loss}
\end{figure}

\begin{figure}[t]
	\centering
    \includegraphics[width=1\linewidth]{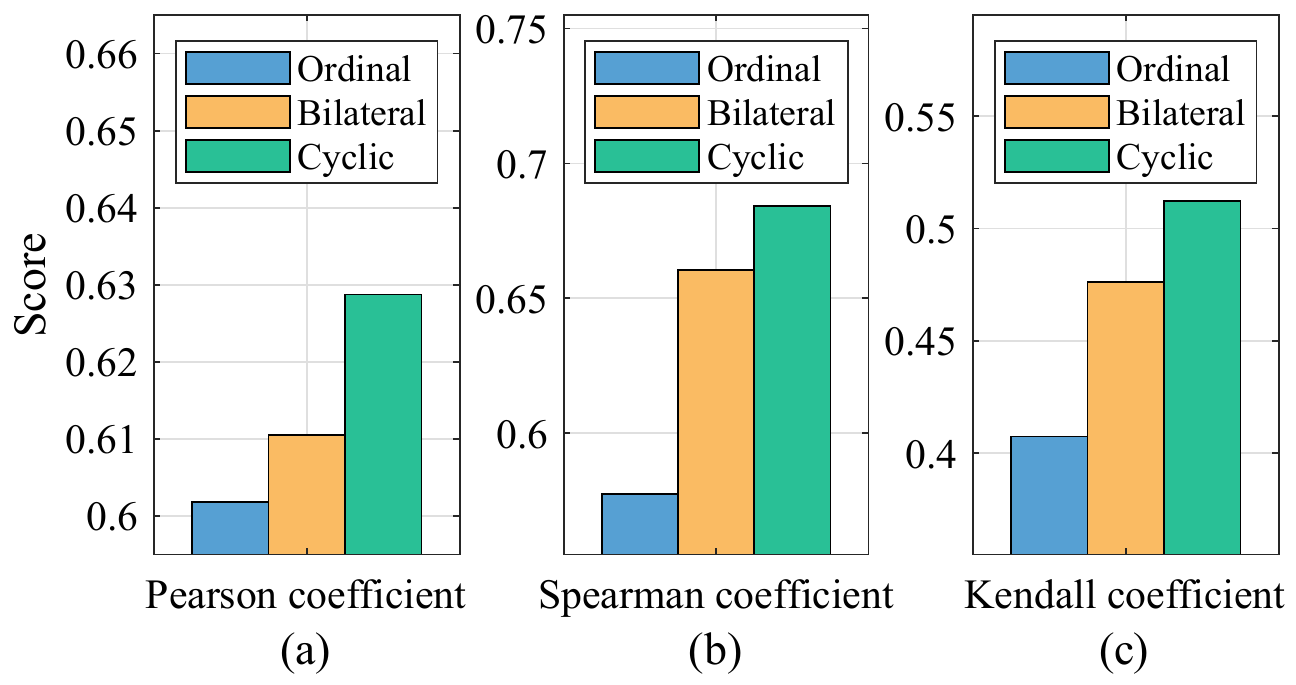}
	\caption{Performance comparison of coefficients~\wrt~ordinal, bilateral, and cyclic mechanisms with $2000$ sampled paths.}
	\label{coefficients}
\end{figure}


\subsection{Ablation Studies} \label{transfer}

\textbf{Effect of ViTAS as a superformer.} To validate the effectiveness of our proposed ViTAS, as in Table~\ref{Ablation_ViTAS}, we implemented the search with $1.4$G FLOPs budget and Twins transformer space on ImageNet-$1$k dataset. Our baseline superformers are AutoSlim and BCNet that adopt ordinal (\#$0$) or bilateral weight (\#$1$) sharing mechanism, respectively, to evaluate a sampled architecture.  Our cyclic pattern (\#$2$) can enjoy a gain of $0.3\%$ or $0.7\%$ on Top-$1$ accuracy compare to bilateral or ordinal pattern, respectively. Moreover, when searching with our proposed weak augmentation strategy, our method (\#$5$) can attains $1.3\%$ or $0.9\%$ performance gain compare to baseline methods (\#$3$~\&~\#$4$).

\textbf{Comparison of ordinal, bilaterally, and cyclic \wrt the superformer training.} As in Figure~\ref{fig:training_loss}, we depict the average of training loss of each epoch~\wrt~three weight sharing mechanisms. In general, two obvious phenomena can be concluded as follows:
\begin{itemize}
    \item In the first few epochs (\eg, $\leq20$), the ordinal superformer has the fastest convergence, then is the bilateral pattern, the last is our cyclic one.
    \item After a few epochs (\eg, $\geq100$), the superformer with the cyclic pattern can be best trained with the lowest loss value, while the bilateral pattern has the second convergence speed, and ordinal pattern performs the worst for training superformer.
\end{itemize}
It is because the ordinal pattern has the largest bias in training channels (as shown in Figure~\ref{fig:training_loss}(a)). A part of the channels converges the fastest in the first few epochs. The bilateral pattern performs similar due to no influence uniformity considered. However, after training more epochs, many channels do not obtain well treated in the superformer of the ordinal and bilateral patterns, thus they present larger average loss values than the cyclic one.

\textbf{Effect of identity shifting strategy.} As in Figure~\ref{fig:training_loss}(b), we present the training losses of the ViTAS using identity shifting strategy and original setting, respectively, on ImageNet-$100$ dataset~\cite{Imagenet,tian2020contrastive}. With the redundancy paths removed, our method can converge to a much smaller loss than the original one, which indicates the proposed identity shifting strategy can promote the training of the superformer. Concretely, the training loss of original and identity shifting decrease to 2.4 and 1.5 at the final, respectively, which indicates that our method promote to better convergence for superformer. Moreover, the results trained from scratch of the searched architectures with identity shifting or original setting is $90.4\%$ and $88.3\%$, respectively.

\begingroup
\setlength{\tabcolsep}{2pt}
\begin{table}[t]
	\caption{Ablation studies of the proposed ViTAS. We implemented the search on the ImageNet-$1$k set with $1.4$G FLOPs budget. Weak\_aug: training the superformer with weak augmentation. Best results are highlighted in bold.}
	\label{Ablation_ViTAS}
	\centering
 	\small{
	\begin{tabular}{c|ccc|c|c} 
	    \hline
		\# & Ordinal & Bilateral & Cyclic & Weak\_aug & Top-1(\%)\\ 
		\hline
		$0$ &\Checkmark & \lightgray{\XSolidBrush} & \lightgray{\XSolidBrush} & \lightgray{\XSolidBrush} & 77.2\\ 
		$1$ & \lightgray{\XSolidBrush} & \Checkmark & \lightgray{\XSolidBrush} & \lightgray{\XSolidBrush} & 77.6\\ 
		$2$ & \lightgray{\XSolidBrush} & \lightgray{\XSolidBrush} & \Checkmark& \lightgray{\XSolidBrush} & 77.9\\ 
		$3$ &\Checkmark & \lightgray{\XSolidBrush} & \lightgray{\XSolidBrush} &\Checkmark & 78.1\\ 
		$4$ & \lightgray{\XSolidBrush} & \Checkmark & \lightgray{\XSolidBrush} &\Checkmark & 78.5\\ 
		$5$ & \lightgray{\XSolidBrush} & \lightgray{\XSolidBrush} & \Checkmark & \Checkmark & $\boldsymbol{79.4}$\\ 
		\hline
	\end{tabular}}
	\vspace{-3mm}
\end{table}
\endgroup

\textbf{Performance comparison~\wrt~ordinal, bilateral, and cyclic weight sharing mechanism with $2000$ sampled paths.} To perform the search, we uniformly assign $8$ budget range from $1$G to $8$G FLOPs, with $250$ paths in each weight sharing mechanism. Generally, we assume architectures' performance are positively correlated with FLOPs. Therefore, we can obtain the scores of the three patterns~\wrt~pearson, spearman, and kendall coefficients~\wrt~different FLOPs groups. As shown in Figure~\ref{coefficients}, our method achieves remarkable improvements comparing to the others, which indicates that our superformer can provide more precisely ranking for architectures. 

\section{Conclusion}
In this paper, we presented a vision transformer architecture search (\ie, ViTAS) framework with the formulated cyclic weight sharing paradigm for the fair ranking of dimensions and also search efficiency. Besides, we propose the identity shifting strategy to arrange the the ID operation at the deeper layers for removing the redundant paths in the superformer. Moreover, we also investigated the training strategy of the superformer and proposed the weak augmentation strategy during search to boost the performance of ViTAS. Extensive experiments on ImageNet-$1$k, COCO$2017$, and ADE$20$k datasets~\wrt~Twins- and DeiT-based transformer space prove the effectiveness of our ViTAS in terms of performance and efficiency.



{\small
\bibliographystyle{ieee_fullname}
\bibliography{egbib}
}

\newpage 
\onecolumn
\appendix

\section{Appendix}
\subsection{Details of Training and Search with ViTAS}

In this section, we present the details of training recipe~\wrt~different models and datasets. Concretely, we uniformly partitioned transformer space with $10$ groups to implement the search of dimensions.

\textbf{Search with ViTAS on ImageNet-$1$k dataset}. We use the same training recipe of superformer for both DeiT- and Twins-based transformer space. As illustrated in Table~\ref{Training_Recipe}, all superformers are trained for $300$ epochs with a batch size of $1,024$ with the AdamW optimizer~\cite{loshchilov2017decoupled}. The learning rate is initialized to be $0.001$ and annealed decayed to zero with a cosine strategy. Besides, we also leverage a linear warm-up in the first five epochs same as~\cite{chu2021twins}. Note that we do not leverage any further extra data augmentation/regularization for training the superformer.

\textbf{Training recipe of searched models on ImageNet-$1$k dataset.} For DeiT-based architectures, we follow the same recipe as~\cite{touvron2020training}. In detail, we use a weight decay of $0.05$ and batch size of $1,024$, and we train the models by $300$ epochs with the learning rate decayed with cosine strategy from initial value $0.05$ to $0$. Except for repeated augmentation, we adopt all other recipes same as DeiT~\cite{touvron2020training},~\ie, $5$ epochs for warmup, $0.1$ of label smoothing, $0.1$ of stochastic depth, rand augmentation, mixup, cutmix, and random erasing. For Twins-based architectures, compare to DeiT recipe, we use the stochastic depth augmentation of $0.1$, $0.2$, $0.3$, $0.5$ for tiny, small, base, and large models, respectively. We also use gradient clipping with a max norm of $5.0$ to stabilize the training process for twins as in~\cite{chu2021twins}.

\textbf{Training recipe of models on COCO$2017$ dataset.} We transfer the searched architectures to COCO$2017$~\cite{COCO} and ADE$20$k~\cite{zhou2017scene} based on MMDetection~\cite{chen2019mmdetection} and with the same recipes as~\cite{chu2021twins}. We train all the models with AdamW optimizer of $12$ epochs and batch size of $16$. The learning rate is initialized as $1\times10^{-4}$ and $2\times10^{-4}$ for RetinaNet and Mask R-CNN, respectively. The learning rate is started with $500$-iteration warmup and decayed by $10\times$ at the $8$-th and $11$-th epoch, respectively. We set stochastic drop path regulation as $0.2$ and weight decay of $0.0001$.

\textbf{Training recipe of models on ADE$20$k dataset.} For the semantic FPN framework, we leverage the same setting as~\cite{wang2021pyramid,chu2021twins}. We train models with AdamW and $80$K steps with a batch size of $16$. The learning rate is set as $1\times10^{-4}$ and decayed with ``poly strategy'' with the power coefficient of $0.9$. We use the drop-path rate of $0.2$ for small, base models while $0.4$ for large model to avoid over-fitting. For the UperNet framework, we leverage the recipe provided in~\cite{xiao2018unified,chu2021twins}. In detail, we train models with AdamW optimizer for $160$ iterations and a batch size of $16$. The initial learning rate is set to $6\times10^{-5}$ and linearly decay to zero. The drop-path is set to $0.2$ for backbone and weight decay is of $0.01$ for the whole model.

\subsection{Transformer Space of ViTAS}

In this section, we present the constitution of transformer space~\wrt~different sizes for the Twins and DeiT. We incorporate all the essential elements in our transformer space, including \emph{head number}, \emph{patch size}, \emph{output dimension of each layer}, and \emph{depth of the architectures}.

Table~\ref{fig:twins_search_space}~describes the defined Twins-based ViTAS transformer space. The block setting (the default order and numbers of MHSA and FC layers) is inspired by~\cite{dosovitskiy2020image,touvron2020training}. The maximum depth of each stage is set as original setting of Twins plus two for ID search and the head number $h$ in the MHSA is chosen within the set $\{3, 6, 12, 16\}$. Meanwhile, the way of sharing FC$_1$ in the MHSA is to evenly divide its output features into $h$ groups (\ie, $h$ heads), while different $h$ leads to different dimensions $D/h$ of a group. Furthermore, in order to accommodate the output dimensions~\wrt~each stage, the maximum dimension of each layer is set to $2\times$ of baseline method, and the dimension can be selected from a group of ten settings,~\ie, $\{\frac{i}{10}\}_{i=1}^{10}$. Given the defined transformer space, we encourage each block in the ViT to freely select their own optimal head numbers and output dimensions. Therefore, with the Twins-small based transformer space as an example, the size of transformer space amounts to $1.1 \times 10^{54}$ and the FLOPs (parameters) ranges from $0.02$G ($0.16$M) to $11.2$G ($86.1$M).

\begin{figure}[H]
	\caption{Macro transformer space for the ViTAS of Twins-based architecture.``TBS'' indicates that layer type is searched from parametric operation and identity operation for depth search. ``Embedding'' represents the patch embedding layer. ``$\text{Max}_\text{a}$'' and ``$\text{Max}_\text{m}$'' indicates the max dimension of attention layer (``$\text{Max}_\text{a}$'' also used for patch embedding layer) and MLP layer, respectively. ``Ratio'' means the reduction ratio from the ``Max Output Dim''. A larger ``Ratio'' indicates a larger dimension.}
    \centering
    \begin{subfigure}[t]{.49\linewidth}
        \centering
    \subcaption{Twins tiny transformer space.}
\resizebox{1\linewidth}{!}{
	\begin{tabular}{cccccccccc} 
	    \hline
		Number & OP & Type & Patch size / \#Heads & $\text{Max}_\text{a}$ & $\text{Max}_\text{m}$ & Ratio \\ 
		\hline
		$1$ & False & Embeding & 4 & $128$ & - & $\{i/10\}_{i=1}^{10}$\\ 
		\multirow{2}*{$4$} & \multirow{2}*{TBS} & Local & \multirow{2}*{$\{2, 4, 8, 16\}$} & \multirow{2}*{$480$} & \multirow{2}*{$512$} & \multirow{2}*{$\{i/10\}_{i=1}^{10}$} \\
		&  & Global & & & & \\ \hline
		$1$ & False & Embeding & 2 & $256$ & - & $\{i/10\}_{i=1}^{10}$\\
		\multirow{2}*{$4$} & \multirow{2}*{TBS} & Local & \multirow{2}*{$\{2, 4, 8, 16\}$} & \multirow{2}*{$960$} & \multirow{2}*{$1024$} & \multirow{2}*{$\{i/10\}_{i=1}^{10}$} \\
		&  & Global & & & & \\ \hline
		$1$ & False & Embeding & 2 & $512$ & - & $\{i/10\}_{i=1}^{10}$\\
		\multirow{2}*{$12$} & \multirow{2}*{TBS} & Local & \multirow{2}*{$\{2, 4, 8, 16\}$} & \multirow{2}*{$1920$} & \multirow{2}*{$2048$} & \multirow{2}*{$\{i/10\}_{i=1}^{10}$} \\
		&  & Global & & & & \\ \hline
		$1$ & False & Embeding & 2 & $1024$ & - & $\{i/10\}_{i=1}^{10}$\\
		\multirow{2}*{$6$} & \multirow{2}*{TBS} & Local & \multirow{2}*{$\{2, 4, 8, 16\}$} & \multirow{2}*{$3840$} & \multirow{2}*{$4096$} & \multirow{2}*{$\{i/10\}_{i=1}^{10}$} \\
		&  & Global & & & & \\
		\hline
	\end{tabular}}
    \end{subfigure}
    \begin{subfigure}[t]{.49\linewidth}
        \centering
    \subcaption{Twins small transformer space.}
        \resizebox{1\linewidth}{!}{
	\begin{tabular}{cccccccccc} 
	    \hline
		Number & OP & Type & Patch size / \#Heads & $\text{Max}_\text{a}$ & $\text{Max}_\text{m}$ & Ratio \\ 
		\hline
		$1$ & False & Embeding & 4 & $128$ & - & $\{i/10\}_{i=1}^{10}$\\ 
		\multirow{2}*{$4$} & \multirow{2}*{TBS} & Local & \multirow{2}*{$\{2, 4, 8, 16\}$} & \multirow{2}*{$480$} & \multirow{2}*{$512$} & \multirow{2}*{$\{i/10\}_{i=1}^{10}$} \\
		&  & Global & & & & \\ \hline
		$1$ & False & Embeding & 2 & $256$ & - & $\{i/10\}_{i=1}^{10}$\\
		\multirow{2}*{$4$} & \multirow{2}*{TBS} & Local & \multirow{2}*{$\{2, 4, 8, 16\}$} & \multirow{2}*{$960$} & \multirow{2}*{$1024$} & \multirow{2}*{$\{i/10\}_{i=1}^{10}$} \\
		&  & Global & & & & \\ \hline
		$1$ & False & Embeding & 2 & $512$ & - & $\{i/10\}_{i=1}^{10}$\\
		\multirow{2}*{$12$} & \multirow{2}*{TBS} & Local & \multirow{2}*{$\{2, 4, 8, 16\}$} & \multirow{2}*{$1920$} & \multirow{2}*{$2048$} & \multirow{2}*{$\{i/10\}_{i=1}^{10}$} \\
		&  & Global & & & & \\ \hline
		$1$ & False & Embeding & 2 & $1024$ & - & $\{i/10\}_{i=1}^{10}$\\
		\multirow{2}*{$6$} & \multirow{2}*{TBS} & Local & \multirow{2}*{$\{2, 4, 8, 16\}$} & \multirow{2}*{$3840$} & \multirow{2}*{$4096$} & \multirow{2}*{$\{i/10\}_{i=1}^{10}$} \\
		&  & Global & & & & \\
		\hline
	\end{tabular}}
	    \vspace{2mm}
    \end{subfigure}

    \begin{subfigure}[t]{.49\linewidth}
        \centering
    \subcaption{Twins base transformer space.}
        \resizebox{1\linewidth}{!}{
	\begin{tabular}{cccccccccc} 
	    \hline
		Number & OP & Type & Patch size / \#Heads & $\text{Max}_\text{a}$ & $\text{Max}_\text{m}$ & Ratio \\ 
		\hline
		$1$ & False & Embeding & 4 & $192$ & - & $\{i/10\}_{i=1}^{10}$\\ 
		\multirow{2}*{$4$} & \multirow{2}*{TBS} & Local & \multirow{2}*{$\{2, 4, 8, 16\}$} & \multirow{2}*{$480$} & \multirow{2}*{$768$} & \multirow{2}*{$\{i/10\}_{i=1}^{10}$} \\
		&  & Global & & & & \\ \hline
		$1$ & False & Embeding & 2 & $384$ & - & $\{i/10\}_{i=1}^{10}$\\
		\multirow{2}*{$4$} & \multirow{2}*{TBS} & Local & \multirow{2}*{$\{2, 4, 8, 16\}$} & \multirow{2}*{$960$} & \multirow{2}*{$1536$} & \multirow{2}*{$\{i/10\}_{i=1}^{10}$} \\
		&  & Global & & & & \\ \hline
		$1$ & False & Embeding & 2 & $768$ & - & $\{i/10\}_{i=1}^{10}$\\
		\multirow{2}*{$20$} & \multirow{2}*{TBS} & Local & \multirow{2}*{$\{2, 4, 8, 16\}$} & \multirow{2}*{$1920$} & \multirow{2}*{$3072$} & \multirow{2}*{$\{i/10\}_{i=1}^{10}$} \\
		&  & Global & & & & \\ \hline
		$1$ & False & Embeding & 2 & $1536$ & - & $\{i/10\}_{i=1}^{10}$\\
		\multirow{2}*{$4$} & \multirow{2}*{TBS} & Local & \multirow{2}*{$\{2, 4, 8, 16\}$} & \multirow{2}*{$3840$} & \multirow{2}*{$6144$} & \multirow{2}*{$\{i/10\}_{i=1}^{10}$} \\
		&  & Global & & & & \\
		\hline
	\end{tabular}}
    \end{subfigure}
    \begin{subfigure}[t]{.49\linewidth}
        \centering
    \subcaption{Twins large transformer space.}
        \resizebox{1\linewidth}{!}{
	\begin{tabular}{cccccccccc} 
	    \hline
		Number & OP & Type & Patch size / \#Heads & $\text{Max}_\text{a}$ & $\text{Max}_\text{m}$ & Ratio \\ 
		\hline
		$1$ & False & Embeding & 4 & $256$ & - & $\{i/10\}_{i=1}^{10}$\\ 
		\multirow{2}*{$4$} & \multirow{2}*{TBS} & Local & \multirow{2}*{$\{2, 4, 8, 16\}$} & \multirow{2}*{$960$} & \multirow{2}*{$1024$} & \multirow{2}*{$\{i/10\}_{i=1}^{10}$} \\
		&  & Global & & & & \\ \hline
		$1$ & False & Embeding & 2 & $512$ & - & $\{i/10\}_{i=1}^{10}$\\
		\multirow{2}*{$4$} & \multirow{2}*{TBS} & Local & \multirow{2}*{$\{2, 4, 8, 16\}$} & \multirow{2}*{$1920$} & \multirow{2}*{$2048$} & \multirow{2}*{$\{i/10\}_{i=1}^{10}$} \\
		&  & Global & & & & \\ \hline
		$1$ & False & Embeding & 2 & $1024$ & - & $\{i/10\}_{i=1}^{10}$\\
		\multirow{2}*{$20$} & \multirow{2}*{TBS} & Local & \multirow{2}*{$\{2, 4, 8, 16\}$} & \multirow{2}*{$3840$} & \multirow{2}*{$4096$} & \multirow{2}*{$\{i/10\}_{i=1}^{10}$} \\
		&  & Global & & & & \\ \hline
		$1$ & False & Embeding & 2 & $2048$ & - & $\{i/10\}_{i=1}^{10}$\\
		\multirow{2}*{$4$} & \multirow{2}*{TBS} & Local & \multirow{2}*{$\{2, 4, 8, 16\}$} & \multirow{2}*{$7680$} & \multirow{2}*{$8192$} & \multirow{2}*{$\{i/10\}_{i=1}^{10}$} \\
		&  & Global & & & & \\
		\hline
	\end{tabular}}
    \end{subfigure}
    \label{fig:twins_search_space}
\end{figure}

Similarly, Table~\ref{fig:deit_search_space} describes the definition of DeiT-based transformer space. ``Max Dim'' indicates the output dimensions of both attention and MLP blocks. With the DeiT-small transformer space, the size of transformer space amounts to $5.4 \times 10^{34}$ and the FLOPs (parameters) ranges from $0.1$G ($0.5$M) to $20.0$G ($97.5$M).

\begin{figure}[H]
	\caption{Macro transformer space for the ViTAS of DeiT-based architecture.``TBS'' indicates that layer type is searched from vanilla ViT block or identity operation for depth search. ``Ratio'' means the reduction ratio from the ``Max Dim''. A larger ``Ratio'' means a larger dimension.}
    \centering
    \begin{subfigure}[t]{.49\linewidth}
        \centering
    \subcaption{DeiT tiny transformer space.}
	\resizebox{1\linewidth}{!}{
	\begin{tabular}{cccccccccc} 
	    \hline
		Number & OP & Type & Patch size / \#Heads & Max Dim & Ratio \\ 
		\hline
		$1$ & False & Linear & $\{14, 16, 32\}$ & $384$ & $\{i/10\}_{i=1}^{10}$\\ 
		\multirow{2}*{$14$} & \multirow{2}*{TBS} & MHSA & $\{3, 6, 12, 16\}$ & $1440$ & $\{i/10\}_{i=1}^{10}$ \\
		&  & MLP & - & $1440$ & $\{i/10\}_{i=1}^{10}$ \\ 
		\hline
	\end{tabular}}
    \end{subfigure}
    \begin{subfigure}[t]{.49\linewidth}
        \centering
    \subcaption{DeiT small transformer space.}
        \resizebox{1\linewidth}{!}{
	\begin{tabular}{cccccccccc} 
	    \hline
		Number & OP & Type & Patch size / \#Heads & Max Dim & Ratio \\ 
		\hline
		$1$ & False & Linear & $\{14, 16, 32\}$ & $768$ & $\{i/10\}_{i=1}^{10}$\\ 
		\multirow{2}*{$14$} & \multirow{2}*{TBS} & MHSA & $\{3, 6, 12, 16\}$ & $2880$ & $\{i/10\}_{i=1}^{10}$ \\
		&  & MLP & - & $2880$ & $\{i/10\}_{i=1}^{10}$ \\ 
		\hline
	\end{tabular}}
	    \vspace{2mm}
    \end{subfigure}
    \label{fig:deit_search_space}
\end{figure}

\subsection{Evolutionary Search}

To avoid the exhausted search from the enormous (\eg, $1.1\times10^{54}$ for Twins small) transformer space and boost the search efficiency, we leverage the multi-objective NSGA-II~\cite{deb2002fast} algorithm for evolutionary search, which is easy to accommodate the constraint budgets (\eg, FLOPs, GPU throughput).  Concretely, we set the population size and generation number as $50$ and $40$, respectively, which amounts to $2,000$ searched paths in ViTAS. To implement the search, we randomly select $50$ paths within the pre-set FLOPs as the initial population. Then, we select the top $20$ performance architectures as the parents to generate new generalization architectures via mutation and crossover. After the search, we only leverage the architecture with the highest performance during search to train from scratch and report its performance. 

\subsection{Coefficient Factors~\wrt~Kendall, Pearson, and Spearman}

In Section~\ref{transfer}, we provide a detailed comparison between ordinal, bilateral, and our cyclic weight sharing paradigm on $2,000$ searched paths~\wrt~three coefficient factors. Indeed, the Pearson $\rho_S$ coefficient aims to evaluate to what degree a monotonic function fits the relationship between two random variables. Besides, the Spearman $\rho_S$ is defined as the Pearson correlation coefficient between the rank variables. Therefore, Pearson and Spearman coefficients share the same formulated equation Eq.~\eqref{eq:appendix:rank:spearman-rho-var} but with different value types (\eg, original value and ranks for Pearson and Spearman coefficients, respectively). Defining $\r$ and $\s$ as two groups of data, the Spearman coefficient $\rho_S$ can be computed by
    \begin{equation}
        \rho_S = \frac{\operatorname{cov}(r, s)}{\sigma_r \sigma_s},
        \label{eq:appendix:rank:spearman-rho-var}
    \end{equation}
where $\operatorname{cov}(\cdot, \cdot)$ is the covariance of two variables, and $\sigma_r$ and $\sigma_s$ are the standard deviations of $r$ and $s$, respectively. In our experiments, the ranks are distinct integers. Therefore, the Eq.~\eqref{eq:appendix:rank:spearman-rho-var} can be also reformulated as
     \begin{equation}
        \rho_S = 1 - \frac{6 \sum_{i=1}^{n} (\boldsymbol{r}_i - \boldsymbol{s}_i)^2}{n (n^2 - 1)},
        \label{eq:appendix:rank:spearman-rho-vector}
    \end{equation}
where $n=2000$ defines the number of overlapped elements between variables.

The Kendall $\tau$ coefficient aims to evaluate the pairwise ranking performance. Given a pair of ($\r_i,\r_j$) and ($\s_i,\s_j$), if we have either both $\r_i>\r_j$ and $\s_i>\s_j$, or both $\r_i<\r_j$ and $\s_i<\s_j$, these two pairs are considered as \textbf{concordant}. Otherwise, it is said to be \textbf{disconcordant}. With the concordant and disconcordant pairs, the Kendall $\tau$ can be formulated as 
\begin{equation}
    K_{\tau} = \frac{n_{\text{con}} - n_{\text{discon}}}{n_{\text{all}}},
    \label{eq:appendix:rank:kendall-tau-var}
\end{equation}
where $n_{\text{con}}$ and $n_{\text{discon}}$ represent the number of concordant and disconcordant pairs, and $n_{\text{all}}$ is the total number of pairs.

\subsection{Ablation Studies of Private Tokens with DeiT-based Superformer} 

For DeiT-based architecture, it usually leverages the trainable vector named class token for to perform the prediction task. The class token is appended to the patch tokens before the first layer and go through the architecture for the classification task. However, for the superformer, different architectures assume to share the weights with the weight sharing paradigm, which blurs the performance gap with the shared class token. Indeed, we propose to private the class tokens to cater for the variance of different paths in superformer. 

To evaluate the effect of private class tokens on superformer, we implement the search with/without private tokens and retrain the searched architectures from scratch and report the performance as in Table~\ref{private_tokens}.

\begingroup
\begin{table}[H]
	\caption{Ablation studies of the proposed ViTAS~\wrt~private token. $\dagger$: architectures that are searched with private token.}
	\label{private_tokens}
	\centering
 	\small{
	\begin{tabular}{c||ccc||cc} 
	    \hline
		models & FLOPs (G) & Throughput (image/s) & Params (M) & Top-$1$(\%) & Top-$5$(\%)\\ 
		\hline
		ViTAS\_DeiT\_A & $1.4$ & $2,842.7$ & $6.8$ & $75.1$ & $92.3$ \\
		ViTAS\_DeiT\_A$\dagger$ & $1.4$ & $2,831.1$ & $6.6$ & $75.6$ & $92.5$ \\
		ViTAS\_DeiT\_B & $5.0$ & $1,134.8$ & $24$ & $79.9$ & $95.0$ \\
        ViTAS\_DeiT\_B$\dagger$ & $4.9$ & $1,189.4$ & $23$ & $80.2$ & $95.1$ \\
		\hline
	\end{tabular}}
\end{table}
\endgroup

\newpage

\subsection{Comparisons between ViTAS and AutoFormer~\cite{chen2021autoformer}} \label{compare_autoformer}

\begin{figure}[H]
    \centering
    
	\begin{subfigure}[t]{1.0\linewidth}
        \centering
        \includegraphics[width=1\linewidth]{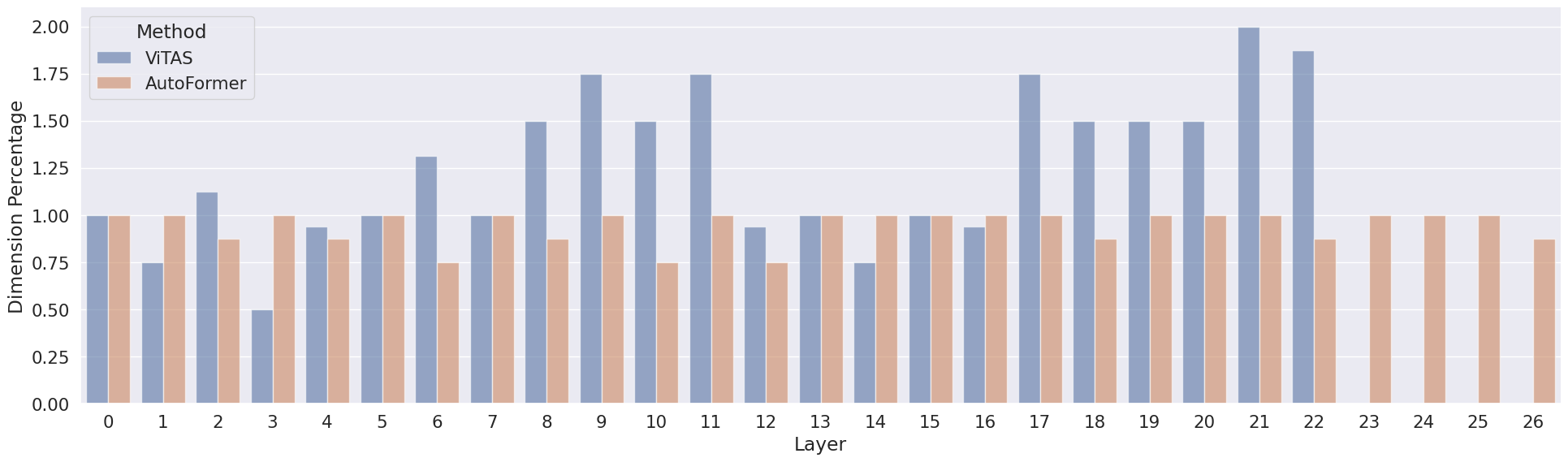}
        \subcaption{Dimension percentage of ViTAS and AutoFormer~\wrt~DeiT small setting.}
    \end{subfigure}
	\begin{subfigure}[t]{1.0\linewidth}
        \centering
        \includegraphics[width=1\linewidth]{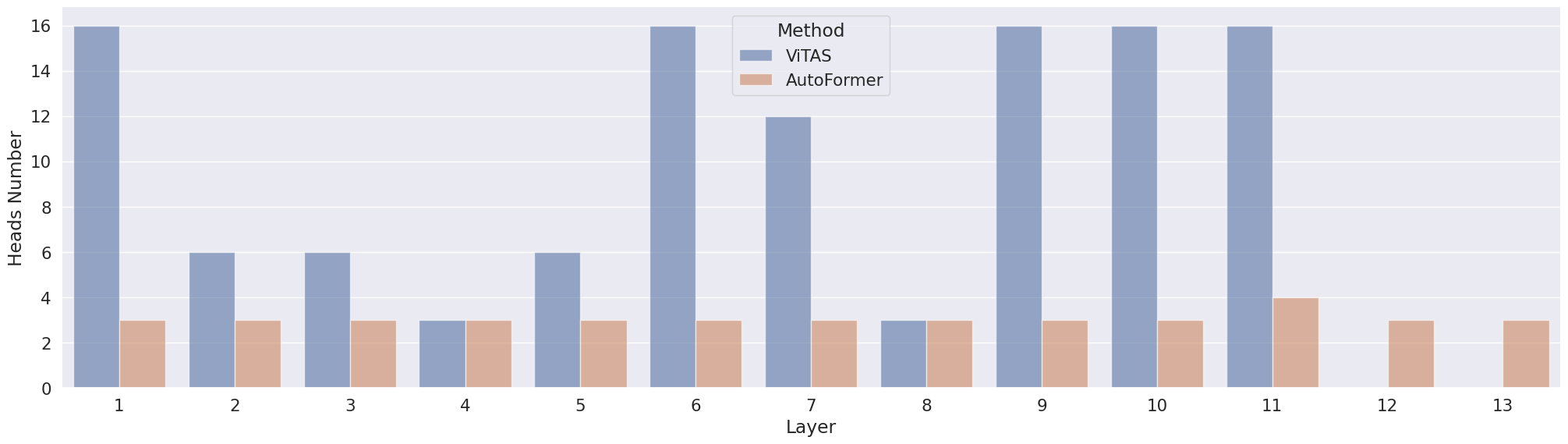}
        \subcaption{Heads number of searched DeiT-T based architecutre~\wrt~ViTAS and AutoFormer.}
    \end{subfigure}
    \caption{Visualization of dimensions and heads number of searched DeiT-T based architecture~\wrt~ViTAS and AutoFormer on ImageNet-$1$k dataset. }\label{fig:Visualization_ViTAS_autoformer}
\end{figure}

To intuitively check the effect of ViTAS with another baseline method,~\ie~AutoFormer, we visualize the network width searched by ViTAS and AutoFormer for $1.4$G FLOPs DeiT-based architecture in Figure~\ref{fig:Visualization_ViTAS_autoformer}. The dimension percentage is computed based on the DeiT-T,~\eg, $1.0$ means that keep the same channels in the corresponding layer (\ie, patch embedding, attention or MLP) as DeiT-T. 

Concretely, in the dimension level (\ie, see Figure~\ref{fig:Visualization_ViTAS_autoformer}(a)), ViTAS keeps smaller dimension in the first few layers while a bit more dimensions in the last few layers. Besides, in the first few layers, the searched ViT architecture tend to keep smaller attention dimension than MLP output dimension, while larger attention dimension in the last few layers. We think this may be because attention is performed based on the extracted information of features, which may be more useful after MLP layer extracted enough information from the input. With tight budget (\ie, $1.4$G), our searched architecture tend to stack less blocks (\ie, $11$) than AutoFormer. In general, AutoFormer keeps almost the same dimensions for all layers as DeiT-T.

When it comes to the heads number, our searched architectures tend to have larger heads number in the first and last few blocks. We think that this may help the architecture to deal with sophisticated information with more heads number in the last few blocks. The AutoFormer keeps almost the same heads number as DeiT-T in all blocks.

\newpage

Moreover, to promote the fair comparison of ViTAS and AutoFormer, we retrain the released structure of AutoFormer with the same retraining recipe of ours, as shown in Table~\ref{tbl:deit_compare}. With the same retraining recipe and same FLOPs budgets, ViTAS achieves $0.8\%$ higher on top-$1$ accuracy than AutoFormer with $1.3$G FLOPs budget DeiT-based architecture, which indicates the effectiveness of our method.

\begin{table}[H]
	\caption{Performance comparison with AutoFormer~\cite{chen2021autoformer} of DeiT tiny based architectures on ImageNet-$1$k by the same training recipe. $\star$ indicates the re-implementation results of important baseline methods with our recipe. $\dagger$: we uniformly scale the searched models to the same FLOPs of $1.3$G~\wrt~AutoFormer-T.}
	\label{tbl:deit_compare}
	\centering
	\setlength\tabcolsep{1.5pt}
	\small
	\begin{tabular}{r||ccc||ccccc} 
	    \hline
		Method & \tabincell{c}{FLOPs\\ (G)} & \tabincell{c}{Throughput\\ (image/s)} &\tabincell{c}{Params \\ (M)} & \tabincell{c}{Top-$1$\\ (\%)}&\tabincell{c}{Top-$5$\\ (\%)}\\ 
		\hline
		DeiT-T$^{\star}$ \cite{touvron2020training} & $1.3$ & $2728.5$ & $5$ & $72.3$ & $91.4$   \\
		AutoFormer-T$^{\star}$ \cite{chen2021autoformer} & $1.3$ & $2955.4$ & $5.7$ & $74.7$ & $91.9$ \\
		ViTAS-DeiT-A$^{\dagger}$ & $1.3$ & $2965.7$ & $6.1$ & $75.5$ & $92.4$ \\
		ViTAS-DeiT-A & $1.4$ & $2831.1$ & $6.6$ & $75.6$ & $92.5$ \\ 
		\hline
	\end{tabular}
	\vspace{-2mm}
\end{table}

\subsection{Implementing the Search with CNN-based Search Space.}

To comprehensively check the effect of cyclic weight sharing mechanism~\wrt~CNN based search space, we implement the dimension search with ResNet$50$ on ImageNet-$1$k dataset. We leverage the same recipe and search space as~\cite{su2021bcnet}. As shown in Table~\ref{Experiment_pruning}, our searched architecture achieves the superior performance~\wrt~baseline methods of DS\footnote{The DS-ResNet is from the paper of ``Dynamic slimmable network'', CVPR 2021 (oral).}, AutoSlim~\cite{autoslim}, and BCNet~\cite{su2021bcnet}, which indicates the effectiveness of the proposed cyclic weight sharing mechanism.

\begin{table}[H]
	\centering
	\caption{Performance comparison of ResNet$50$ on ImageNet-$1$k. $\star$ indicates the re-implementation results of important baseline methods with our recipe.}
	\label{Experiment_pruning}
	\small
	\resizebox{\textwidth}{!}{
	\begin{tabular}{c|c|ccc|c|c|ccc}
			\hline
			Groups&Methods&FLOPs(G)&Params(M)&accuracy(\%)&Groups&Methods&FLOPs(G)&Params(M)&accuracy(\%)  \\ 
			\hline
			\multirow{4}*{$1.4$G} & DS-ResNet-S & $1.2$ & - & $74.6$ & \multirow{4}*{$2.4$G} & DS-ResNet-M & $2.2$ & - & $76.1$ \\
			& AutoSlim$^{\star}$  & $1.4$ & $15.3$ & $73.8$ && AutoSlim$^{\star}$ & $2.4$ & $21.8$ & $75.7$ \\
			& BCNet$^{\star}$ & $1.4$ & $16.3$ & $75.4$ && BCNet$^{\star}$ & $2.4$ & $22.6$ & $77.0$\\
			& ViTAS & $1.4$ & $15.7$ & $75.8$ && ViTAS & $2.4$ & $22.1$ & $77.2$\\
			\hline
	\end{tabular}}
\end{table}

\subsection{Limitation}

In this paper, we introduce the cyclic weight sharing paradigm for searching the optimal dimensions. Although we have mathematically formulated it in Section~\ref{cyclic_channels}, we believe that there will be more in-depth insights with this clue for the followers, since there no other paper has ever discussed the channel influence issue. For example, rethinking the problem formulation from the aspect of optimal transport in graph theory and redefining the computation technique of the dimension influence in practice.

\subsection{Re-implementation Results of Baseline Methods}
To intuitively check the performance of ViTAS, we present the re-implement results~\wrt~baseline methods on ImageNet-$1$k, COCO$2017$, and ADE$20$k datasets with our training recipe, as shown in Table~\ref{tbl:reimplement}$\sim$\ref{tbl:ADE$20$k_reimplement}.

\begin{table}[H]
	\caption{Searched Twins-based ViT architectures~\wrt~different FLOPs and GPU throughput on ImageNet-$1$k. We abbreviate the name of tiny, short, base, and large for T, S, B, and L, respectively. Re-Top-$1$, Re-Top-$5$: indicates that the re-implementation results of baseline methods with our recipe. Top-$1$, Top-$5$: performance of baseline methods that is reported from papers. Our results are highlighted in bold.} 
	\label{tbl:reimplement}
	\centering
	\small
	\vspace{-2mm}
	\setlength\tabcolsep{1.5pt}
	\begin{tabular}{r||ccc||cc||cc} 
	    \hline
		Method & \tabincell{c}{FLOPs\\ (G)} & \tabincell{c}{Throughput\\ (image/s)} &\tabincell{c}{Params \\ (M)} & \tabincell{c}{Top-$1$\\ (\%)}&\tabincell{c}{Top-$5$\\ (\%)}& \tabincell{c}{Re-Top-$1$\\ (\%)}&\tabincell{c}{Re-Top-$5$\\ (\%)}\\ \hline
		\hline
		DeiT-T~\cite{touvron2020training} & $1.3$ & $2728.5$ & $5$ & 72.2 & 91.3 & $72.3$ & $91.4$   \\
        Twins-T~\cite{chu2021twins} & $1.4$ & $1580.7$ & $11.5$& - & - & $77.8$ & $94.1$ \\
		AutoFormer-T~\cite{chen2021autoformer} & $1.3$ & $3055.4$ & $5.7$ & 74.7 & 92.6 & $74.7$ & $91.9$ \\
		ViTAS-Twins-T & $1.4$ & $1686.3$ & $13.8$ & $\textbf{79.4}$ & $\textbf{94.8}$ & - & - \\  
		\hline
		DeiT-S~\cite{touvron2020training} & $4.6$ & $437.0$ & $22.1$ & 79.8 & - & $79.9$ & $95.0$ \\
		Twins-SVT-S~\cite{chu2021twins} & $2.9$ & $1059$ & $24$ & 81.7 & - & $81.6$ & $95.9$ \\
 		AutoFormer-S~\cite{chen2021autoformer} & $5.1$ & $1231.7$ & $22.9$ & 81.7 & 95.7 & $79.8$ & $95.0$ \\ 
		Twins-PCPVT-S~\cite{chu2021twins} & $3.8$ & $815$ & $24.1$ & 81.2 & - & $81.2$ & $95.6$ \\
		Swin-T~\cite{liu2021swin} & $4.5$ & $766$ & $29$ & 81.3 & - & $81.2$ & $95.5$ \\
		ViTAS-Twins-S & $3.0$ & $958.6$ & $30.5$ & $\textbf{82.0}$ & $\textbf{95.7}$ & - & - \\ 
		\hline
		Swin-S~\cite{liu2021swin} & $8.7$ & $444$ & $50$ & 83.0 & - & $83.0$ & $96.2$ \\
		Twins-SVT-B~\cite{chu2021twins} & $8.6$ & $469$ & $56$ & 83.2 & - & $83.2$ & $96.3$\\
		ViTAS-Twins-B & $8.8$ & $362.7$ & $66.0$ & $\textbf{83.5}$ & $\textbf{96.5}$ & - & - \\  
		\hline
		DeiT-B~\cite{touvron2020training} & $17.6$ & $292$ & $86.6$ & 81.8 & - & $81.8$ & $95.7$ \\
		Twins-SVT-L~\cite{chu2021twins} & 15.1 & 288 & 99.2 & 83.7 & - & 83.6 & 96.6 \\
		ViTAS-Twins-L & $16.1$ & $260.7$ & $124.8$ & $\textbf{84.0}$ & $\textbf{96.9}$ & - & - \\  
		\hline
	\end{tabular}
	\vspace{-3mm}
\end{table}

\begin{table}[H]
	\caption{Object detection and instance segmentation performance with searched backbones on the COCO$2017$ dataset with Mask R-CNN framework and RatinaNet framework. We followed the same training and evaluation setting as~\cite{chu2021twins}. ``FLOPs'' and ``Param'' are in giga and million, respectively. $\star$ indicates the re-implementation results of important baseline methods with our recipe. Methods without $\star$ indicates that the performance is reported from papers. Since Twins do not provide the experiment results for Twins-SVT-L, we only report our re-implement results as Twins-SVT-L$^{\star}$. Our results are highlighted in bold.} 
	\label{tbl:maskrcnn_reimplement}
	\centering
	\setlength\tabcolsep{2pt}
	\small
	\vspace{-3mm}
	\begin{tabular}{r||cc||cccccc||cc||cccccc} 
	    \hline
	    \multirow{2}*{Backbone} & \multicolumn{8}{c||}{Mask R-CNN $1\times$~\cite{he2017mask}} & \multicolumn{8}{c}{RetinaNet $1\times$~\cite{lin2017focal}} \\\cline{2-17}
	    & FLOPs & Param & $\text{AP}^{\text{b}}$ & AP$_{50}^{\text{b}}$ & AP$_{75}^{\text{b}}$ & AP$^{\text{m}}$ & AP$_{50}^{\text{m}}$ & AP$_{75}^{\text{m}}$ & FLOPs & Param & $\text{AP}^{\text{b}}$ & AP$_{50}^{\text{b}}$ & AP$_{75}^{\text{b}}$ & AP$_{S}$ & AP$_{M}$ & AP$_{L}$ \\ 
	    \hline
        Twins-SVT-S~\cite{chu2021twins} & $164$ & $44.0$ & $43.4$ & $66.0$ & $47.3$ & $40.3$ & $63.2$ & $43.4$ & $104$ & $34.3$ & $43.0$ & $64.2$ & $46.3$ & $28.0$ & $46.4$ & $57.5$ \\
        Twins-SVT-S$^{\star}$ & $164$ & $44.0$ & $43.5$ & $66.0$ & $47.8$ & $40.1$ & $62.9$ & $43.1$ & $104$ & $34.3$ & $42.2$ & $63.3$ & $44.9$ & $26.4$ & $45.6$ & $57.0$ \\
        ViTAS-Twins-S & 168 & 44.2 & $\textbf{45.9}$ & $\textbf{67.8}$ & $\textbf{50.3}$ & $\textbf{41.5}$ & $\textbf{64.7}$ & $\textbf{45.0}$ & $108$ & $41.3$ & $\textbf{44.4}$ & $\textbf{65.3}$ & $\textbf{47.6}$ & $\textbf{27.5}$ & $\textbf{48.3}$ & $\textbf{60.0}$ \\ 
        \hline
        Twins-SVT-B~\cite{chu2021twins} & $224$ & $76.3$ & $45.2$ & $67.6$ & $49.3$ & $41.5$ & $64.5$ & $44.8$ & $163$ & $67.0$ & $45.3$ & $66.7$ & $48.1$ & $28.5$ & $48.9$ & $60.5$ \\
        Twins-SVT-B$^{\star}$ & $224$ & $76.3$ & $45.5$ & $67.4$ & $50.0$ & $41.4$ & $64.5$ & $44.5$ & $163$ & $67.0$ & $44.4$ & $65.6$ & $47.4$ & $28.5$ & $47.9$ & $59.5$ \\
        ViTAS-Twins-B & 227 & 85.4 & $\textbf{47.6}$ & $\textbf{69.2}$ & $\textbf{52.2}$ & $\textbf{42.9}$ & $\textbf{66.3}$ & $\textbf{46.5}$ & $167$ & $76.2$ & $\textbf{46.0}$ & $\textbf{66.7}$ & $\textbf{49.6}$ & $\textbf{29.1}$ & $\textbf{50.2}$ & $\textbf{62.0}$ \\ 
        \hline
        Twins-SVT-L$^{\star}$~\cite{chu2021twins} & 292 & 119.7 & 45.9 & 67.9 & 49.9 & 41.6 & 65.0 & 45.0 & 232 & 110.9 & 45.2 & 66.6 & 48.4 & 29.0 & 48.6 & 60.9 \\  
        ViTAS-Twins-L & 301 & 144.1 & $\textbf{48.2}$ & $\textbf{69.9}$ & $\textbf{52.9}$ & $\textbf{43.3}$ & $\textbf{66.9}$ & $\textbf{46.7}$ & 246 & 135.5 & \textbf{47.0} & \textbf{67.8} & \textbf{50.3} & \textbf{29.6} & \textbf{50.9} & \textbf{62.4} \\ 
        \hline
	\end{tabular}
	\vspace{-1mm}
\end{table}

\begin{table}[H]
	\caption{Performance comparisons with searched backbones on ADE$20$K validation dataset. Architectures were implemented with the same training recipe as~\cite{chu2021twins}. All backbones were pretrained on ImageNet-$1$k, except for SETR, which was pretrained on ImageNet-$21$k dataset. $\star$ indicates the re-implementation results of important baseline methods with our recipe. Methods without $\star$ indicates that the performance is reported from papers. mAcc: mean accuracy for all categories. aAcc: accuracy of all pixels. Our results are highlighted in bold.} 
	\label{tbl:ADE$20$k_reimplement}
	\centering
	\setlength\tabcolsep{2pt}
	\small
	\vspace{-2mm}
	\begin{tabular}{r||cc||ccc||cc||ccc} 
	    \hline
	    \multirow{2}*{Backbone} & \multicolumn{5}{c||}{Semantic FPN $80$k~\cite{chu2021twins}} & \multicolumn{5}{c}{Upernet $160$k~\cite{liu2021swin}} \\ \cline{2-11}
	    & FLOPs(G) & Param(M) & mIoU & mAcc & aAcc & FLOPs(G) & Param(M) & mIoU & mAcc & aAcc \\ 
	    \hline
        Twins-SVT-S~\cite{chu2021twins} & $37$ & $28.3$ & $43.2$ & - & - & $228$ & $54.4$ & $46.2$ & - & - \\
        Twins-SVT-S$^{\star}$ & $37$ & $28.3$ & $43.6$ & 55.4 & 80.6 & $228$ & $54.4$ & $45.9$ & 57.3 & 81.5 \\
        ViTAS-Twins-S & 38 & 35.1 & $\textbf{46.6}$ & \textbf{57.6} & \textbf{82.2} & 229 & 61.7 & $\textbf{47.9}$ & \textbf{59.0} & \textbf{82.6} \\
        \hline
        Twin-SVT-B~\cite{chu2021twins} & $67$ & $60.4$ & $45.3$ & - & - & $261$ & $88.5$ & $47.7$ & - & - \\ 
        Twin-SVT-B$^{\star}$ & $67$ & $60.4$ & $45.5$ & 57.0 & 81.5 & $261$ & $88.5$ & $47.7$ & 59.1 & 82.7 \\ 
        ViTAS-Twins-B & 67 & 69.6 & $\textbf{49.5}$ & \textbf{60.5} & \textbf{83.4} & 261 & 97.7 & $\textbf{50.2}$ & \textbf{61.1} & \textbf{83.5} \\
        \hline
        Twins-SVT-L~\cite{chu2021twins} & $102$ & $103.7$ & $46.7$ & - & - & $297$ &  $133$ & $48.8$ & - & - \\ 
        Twins-SVT-L$^{\star}$ & $102$ & $103.7$ & $46.9$ & 58.3 & 81.8 & 297 & $133$ & $49.2$ & 60.5 & 82.8 \\ 
        ViTAS-Twins-L & 108 & 128.2 & $\textbf{50.4}$ & \textbf{61.6} & \textbf{83.6} & 303 & 158.7 & $\textbf{51.3}$ & \textbf{61.9} & \textbf{84.4} \\
	    \hline
	\end{tabular}
	\vspace{-3mm}
\end{table}

\subsection{Comparisons of ViTAS with Baseline Methods~\wrt~Transferability to COCO$2017$ and ADE$20$k Datasets}

\begin{figure}[H]
    \centering
    
	\begin{subfigure}[t]{.48\linewidth}
        \centering
        \includegraphics[height = 0.8\linewidth,width=1\linewidth]{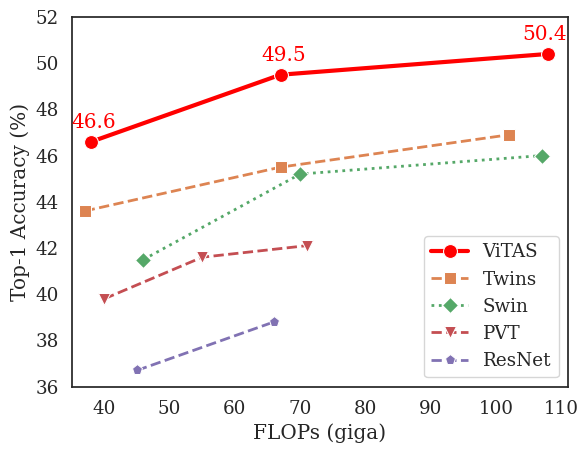}
        \subcaption{Semantic FPN $80$k framework on ADE$20$k dataset.}
    \end{subfigure}
	\begin{subfigure}[t]{.48\linewidth}
        \centering
        \includegraphics[height = 0.8\linewidth,width=1\linewidth]{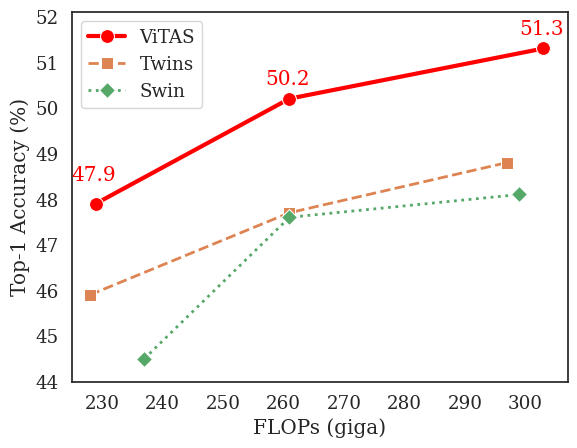}
        \subcaption{Upernet $160$k framework on ADE$20$k dataset.}
    \end{subfigure}
	\begin{subfigure}[t]{.48\linewidth}
        \centering
        \includegraphics[height = 0.8\linewidth,width=1\linewidth]{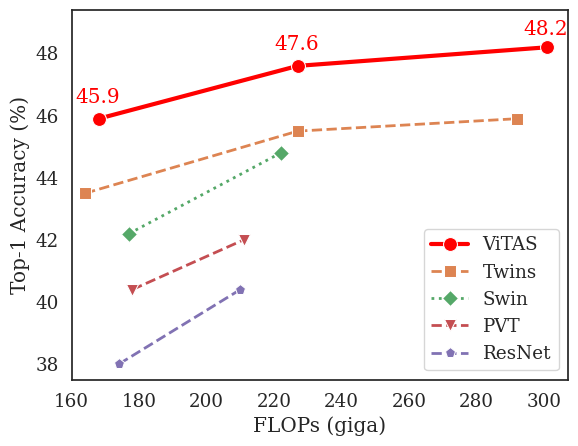}
        \subcaption{Mask R-CNN framework on COCO$2017$ dataset.}
    \end{subfigure}
	\begin{subfigure}[t]{.48\linewidth}
        \centering
        \includegraphics[height = 0.8\linewidth,width=1\linewidth]{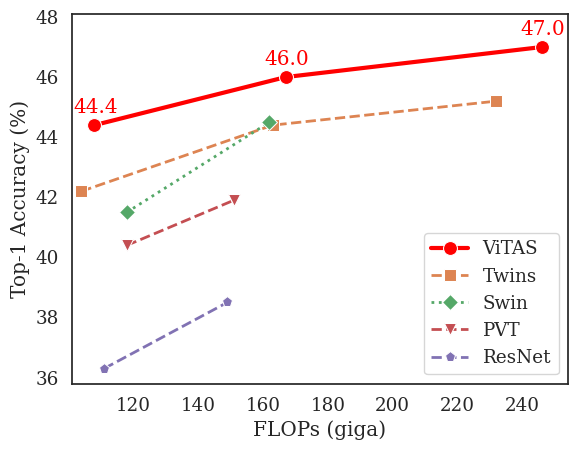}
        \subcaption{RetinaNet framework on COCO$2017$ dataset.}
    \end{subfigure}
    \caption{The comparisons between ViTAS and other baseline methods with on ADE$20$k and COCO$2017$ dataset.}\label{fig:sota_performance}
\end{figure}

To intuitively compare the transferability of ViTAS with other baseline methods, we evaluated the generalization ability of the ViTAS by transferring the searched architectures to COCO$2017$ and ADE$20$k datasets. As shown in Figure~\ref{fig:sota_performance}, we visualize the performance of ViTAS~\wrt~other baseline methods. Indeed, with similar FLOPs budget, our ViTAS achieves the superiority than other baseline methods on the both datasets~\wrt~the tasks of segmentation and detection.

\newpage

\begin{figure}[H]
    \vspace{-2mm}
    \centering
    \includegraphics[width=\linewidth]{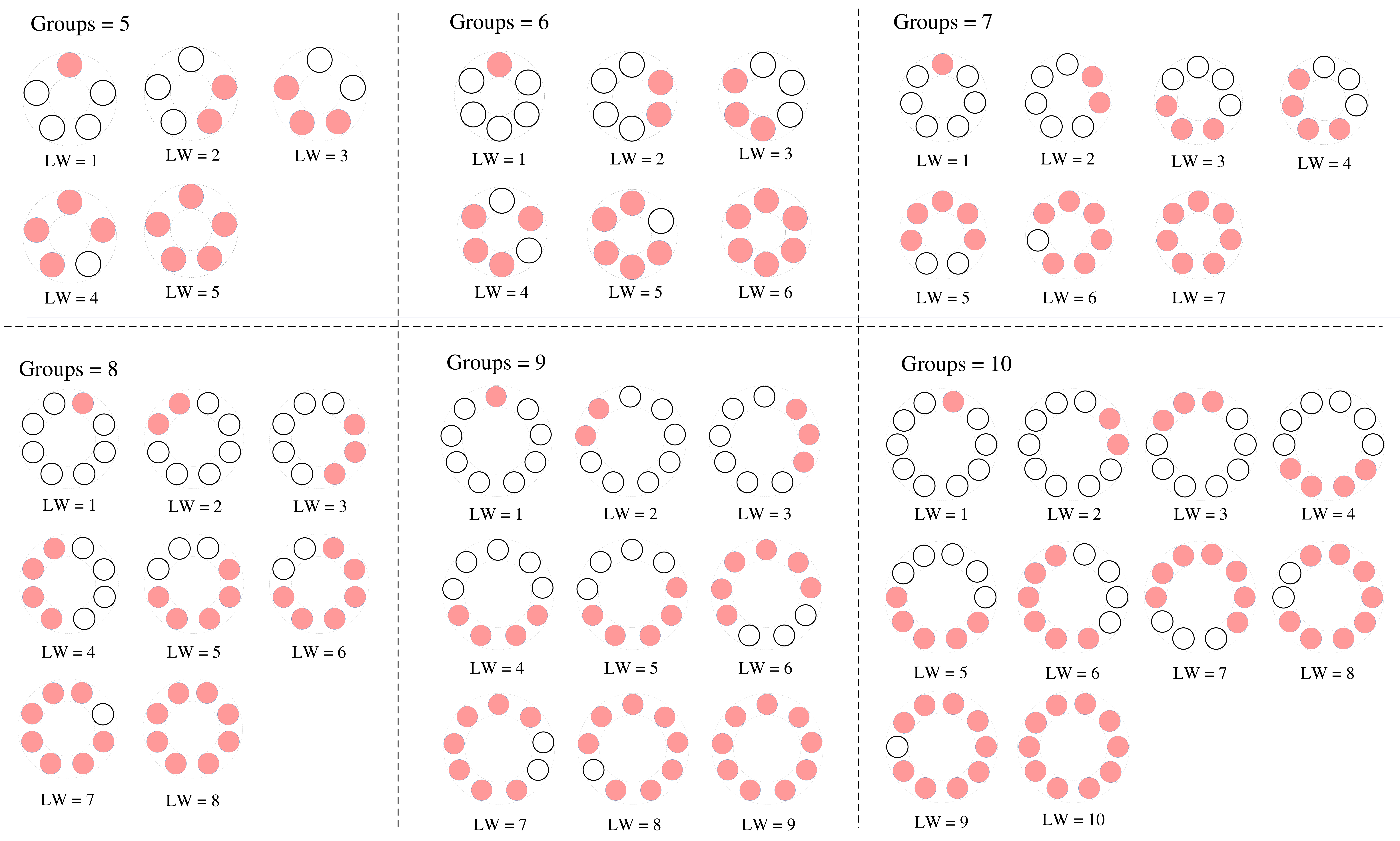}
    \caption{Visualization of cyclic weight sharing mechanism~\wrt~groups number from $5$ to $10$.}\label{fig:cyclic_weight_sharing}
    \vspace{-4mm}
\end{figure}

\subsection{Visualization of Cyclic Weight Sharing Mechanism~\wrt~Different Groups Number}

In this section, we visualize the channels~\wrt~different groups number in cyclic weight sharing mechanism. Indeed, to promote the usage of cyclic weight sharing mechanism, we constraint the channels to be continuous~\wrt~each group, as illustrated in Figure~\ref{fig:cyclic_weight_sharing}.

Moreover, with Eq.~\eqref{eq_objective}, we compare the \textit{influence uniformity} of our cyclic pattern with ordinal and bilateral weight sharing mechanism. As shown in Table~\ref{influence_gap}, our cyclic pattern achieves almost zero influence gap~\wrt~channels and much smaller than others, and bilateral pattern has about the half influence gap than ordinal pattern. It indicates that our ViTAS can fairly train all channels in superformer, and thus can perform better than baseline methods.
\begin{table}[H]
	\caption{Comparison of cyclic weight sharing mechanism with other baseline methods~\wrt~different groups number.}
	\label{influence_gap}
	\centering
 	\vspace{-4mm}
 	\small{
	\begin{tabular}{c||cccccc} 
	    \hline
		\diagbox{methods}{Groups} & $5$ & $6$ & $7$ & $8$ & $9$ & $10$ \\ 
		\hline
		Ordinal & $30.3$ & $48.1$ & $70.6$ & $99.1$ & $130.0$ & $167.4$ \\
		Bilateral & $10.0$ & $19.0$ & $30.6$ & $46.1$ & $66.1$ & $88.7$ \\
        Cyclic & $0.6$ & $0.3$ & $0.7$ & $0.5$ & $0.8$ & $0.5$ \\
		\hline
	\end{tabular}}
 	\vspace{-3mm}
\end{table}


\subsection{Visualization and Interpretation of Searched ViT Architectures}

\begin{figure}[H]
    \centering
    \begin{subfigure}[t]{.47\linewidth}
        \centering
        \includegraphics[width=1\linewidth]{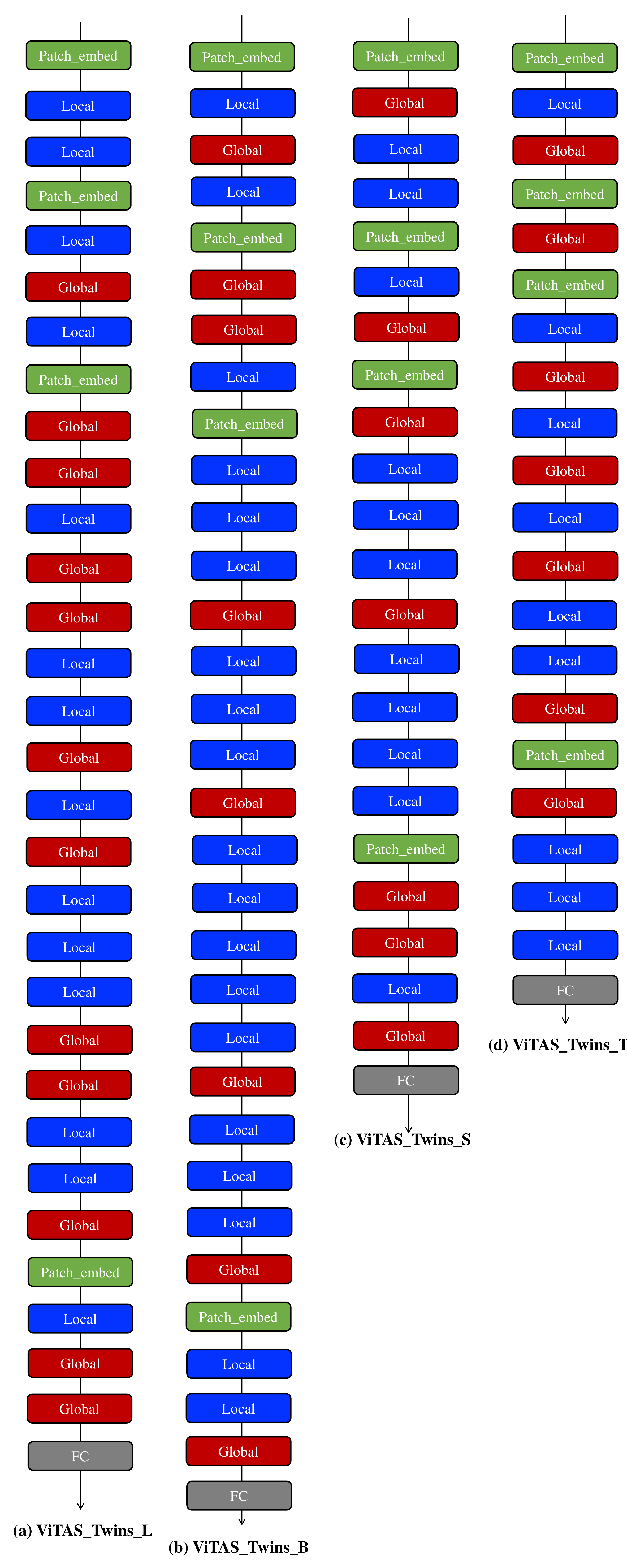}
        \subcaption{Visualization of operations~\wrt~searched architectures.}
    \end{subfigure}
    \begin{subfigure}[t]{.49\linewidth}
        \centering
        \includegraphics[width=1.0\linewidth]{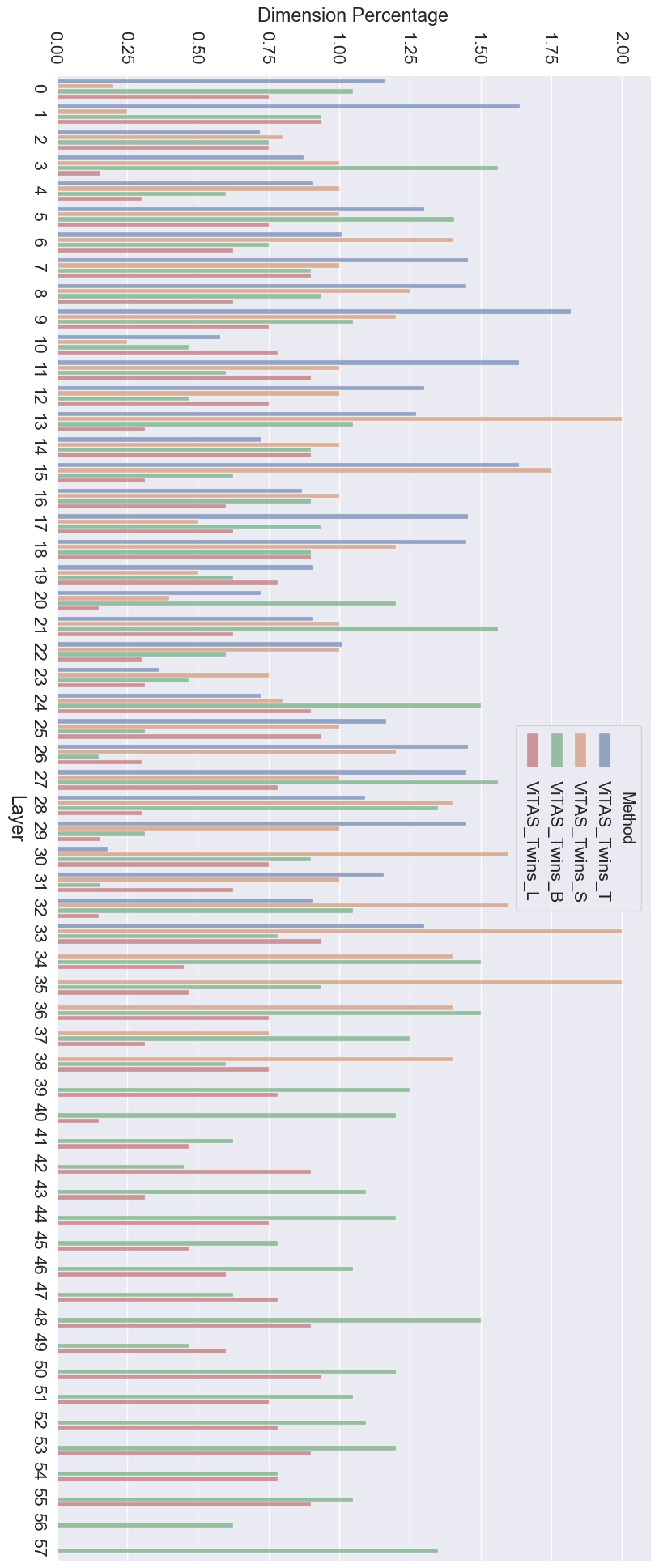}
        \subcaption{Visualization of dimensions~\wrt~searched architectures.}
    \end{subfigure}
    \caption{Visualization of searched architectures on Twins transformer space.}
    \label{fig:ViTAS_visual}
\end{figure}

In this section, we discuss the searched ViT architectures. For intuitively understanding, we visualize our searched three ViT architectures with various FLOPs in Figure~\ref{fig:ViTAS_visual} as examples. From Figure~\ref{fig:ViTAS_visual}, we summarize the three experiential results for further ViT design. For the last stage of ViT architectures, the downsampling size is equal to the feature size, which indicates that ``Local'' blocks perform similarly to ``Global'' blocks.

\begin{itemize}

    \item The optimal architecture generally tends to follow several local operations after the global blocks.
    
    \item The optimal architecture has a bit more local operations than the global operations.
    
    \item The dimension between layers changes smaller in Twins-based architectures than in DeiT-based architectures (\ie, see Figure~\ref{fig:Visualization_ViTAS_autoformer}), which indicates that the work in~\cite{chu2021twins} performs as a strong baseline~\wrt~the provided ViT architectures.
\end{itemize}

\end{document}